\documentclass[10pt,journal,compsoc]{IEEEtran}

\usepackage[ruled, vlined]{algorithm2e}
\usepackage{algpseudocode}
\usepackage{amsmath}
\usepackage{fontawesome}
\usepackage{diagbox}
\usepackage{graphics}
\usepackage{subfigure}
\usepackage{epsfig}
\usepackage{threeparttable}
\usepackage{xspace}
\usepackage{graphicx}
\usepackage{cancel}
\newcommand\Ccancel[2][black]{
    \let\OldcancelColor\CancelColor
    \renewcommand\CancelColor{\color{#1}}
    \cancel{#2}
    \renewcommand\CancelColor{\OldcancelColor}
}
\usepackage{threeparttable}
\usepackage{color}
\usepackage[normalem]{ulem}
\usepackage{multirow}
\usepackage{float}
\usepackage{amsfonts}
\usepackage{bm}
\usepackage{array}
\usepackage[table]{xcolor}
\usepackage{colortbl}
\usepackage{rotating}
\usepackage{booktabs}
\usepackage{overpic}
\usepackage{contour}
\usepackage{enumitem}
\usepackage{cite}
\usepackage{tcolorbox}
\usepackage{amssymb}
\usepackage{pifont}
\usepackage{ragged2e}

\tcbuselibrary{theorems}
\tcbset{highlight math/.append style={left=0mm,right=0mm,top=0mm,bottom=0mm, colframe=white}}

\usepackage{amssymb}
\usepackage{mathtools}
\usepackage{amsthm}

\theoremstyle{plain}
\newtheorem{theorem}{Theorem}[section]

\definecolor{lightgray}{gray}{.9}
\definecolor{deepgray}{gray}{.8}

\definecolor{mygray}{gray}{.9}
\newcolumntype{I}{!{\vrule width 1pt}}

\makeatletter
\newcommand{\thickhline}{%
    \noalign {\ifnum 0=`}\fi \hrule height 1pt
    \futurelet \reserved@a \@xhline
}
\makeatother

\makeatletter
\DeclareRobustCommand\onedot{\futurelet\@let@token\@onedot}
\def\@onedot{\ifx\@let@token.\else.\null\fi\xspace}

\definecolor{mygray}{gray}{.9}
\definecolor{mygreen}{RGB}{93,173,85}
\definecolor{mywarning}{RGB}{233,144,61}
\newcommand{\hlg}[1]{\textcolor{mygreen}{#1}}
\newcommand{\hlr}[1]{\textcolor{red}{#1}} 
\newcommand{\pub}[1]{{\color{gray}{\tiny{[{#1}]}}}}

\usepackage[pagebackref,breaklinks,colorlinks]{hyperref}
\newcommand{\tmark}{\ding{51}}%
\newcommand{\xmark}{\ding{55}}%
\usepackage[capitalize]{cleveref}
\Crefname{table}{Table}{Tables}
\crefname{table}{Tab.}{Tabs.}
\crefname{section}{§}{§§}

\def\eg{\emph{e.g}\onedot} 
\def\ie{\emph{i.e}\onedot}

\newcommand{\solo}{{BASE}}
\newcommand{\fedavg}{{FedAvg}}
\newcommand{\fedprox}{{FedProx}}

\newcommand{\moon}{{MOON}}
\newcommand{\fedproc}{{FedProc}}

\newcommand{\fccl}{{FCCL}}

\newcommand{\fedmd}{{FedMD}}

\newcommand{\feddf}{{FedDF}}
\newcommand{\rhfl}{{RHFL}}

\newcommand{\fedrs}{{FedRS}}

\newcommand{\fedmatch}{{FedMatch}}

\newcommand{\oursabbrv}{{FCCL+}}

\newcommand{\FISL}{Federated Instance Similarity Learning}
\newcommand{\FISLab}{{FISL}}
\newcommand{\FCCM}{Federated Cross-Correlation Matrix}
\newcommand{\FCCMab}{{FCCM}}

\newcommand{\FNTD}{Federated Non Target Distillation}
\newcommand{\FNTDab}{{FNTD}} 

\newcommand{\digits}{{Digits}}
\newcommand{\officehome}{{Office-Home}}
\newcommand{\officeTO}{{Office31}}
\newcommand{\officecaltech}{{Office Caltech}}
\newcommand{\cifarhun}{{Cifar-100}}
\newcommand{\tyimagenet}{{Tiny-ImageNet}}
\newcommand{\fashionmnist}{{Fashion-MNIST}}

\newcommand{\mnist}{{MNIST}}
\newcommand{\usps}{{USPS}}
\newcommand{\svhn}{{SVHN}}
\newcommand{\syn}{{SYN}}

\newcommand{\resnet}{{ResNet}}
\newcommand{\resnetten}{{ResNet10}}
\newcommand{\resnettwelve}{{ResNet12}}
\newcommand{\resneteighteen}{{ResNet18}}
\newcommand{\resnetthirtyfour}{{ResNet34}}
\newcommand{\efficientnet}{{EfficientNet}}
\newcommand{\mobilenet}{{MobileNet}}
\newcommand{\googlenet}{{GoogLeNet}}

\begin{document}

\title{Generalizable Heterogeneous Federated Cross-Correlation and Instance Similarity Learning}

\author{Wenke Huang, Mang Ye,~\IEEEmembership{Senior Member,~IEEE}, Zekun Shi, Bo Du,~\IEEEmembership{Senior Member,~IEEE} 

\IEEEcompsocitemizethanks{
\IEEEcompsocthanksitem 
W. Huang, M. Ye, B. Du and Z. Shi are with the School of Computer Science, Wuhan University, and Hubei Luojia Laboratory, Wuhan, China 430072. \protect
E-mail:\{wenkehuang, yemang, dubo\}@whu.edu.cn and  zekunshi99@gmail.com

\IEEEcompsocthanksitem A preliminary version of this work has appeared in CVPR 2022 {\cite{FCCL_CVPR22}}.

}
}

\markboth{Generalizable Heterogeneous Federated Cross-Correlation and Instance Similarity
Learning}%
{Shell \MakeLowercase{\textit{et al.}}: Bare Demo of IEEEtran.cls for Journals}

\IEEEtitleabstractindextext{
\begin{abstract}
\justifying
Federated learning is an important privacy-preserving multi-party learning paradigm, involving collaborative learning with others and local updating on private data.  Model heterogeneity and  catastrophic forgetting are two crucial challenges, which greatly limit the applicability and generalizability.
  This paper presents a novel \oursabbrv{}, federated correlation and similarity learning with non-target distillation, facilitating the both intra-domain discriminability and inter-domain generalization. 
  For heterogeneity issue, we leverage irrelevant unlabeled public data for communication between the heterogeneous participants. We construct cross-correlation matrix and align instance similarity distribution on both logits and feature levels, which effectively overcomes the communication barrier and improves the generalizable ability. 
  For catastrophic forgetting in local updating stage, \oursabbrv{} introduces \FNTD{}, which retains inter-domain knowledge while avoiding the optimization conflict issue, fulling distilling privileged inter-domain information through depicting posterior classes relation. 
  Considering that there is no standard benchmark for evaluating existing heterogeneous federated learning under the same setting, we present a comprehensive benchmark with extensive representative methods under four domain shift scenarios, supporting both heterogeneous and homogeneous federated settings. Empirical results demonstrate the superiority of our method and the efficiency of modules on various scenarios. The benchmark code for reproducing our results is available at \url{https://github.com/WenkeHuang/FCCL}.
\end{abstract}
}
\maketitle
\IEEEdisplaynontitleabstractindextext
\IEEEpeerreviewmaketitle
\newcommand{\tworow}[2]{
\begin{tabular}{@{}c@{}}
{#1}  \\
{(#2 $\! \%$)}
\end{tabular}
}

\newcommand{\simplecnn}{{SimpleCNN}}

\definecolor{black}{rgb}{0.0, 0.5, 1.0}
\definecolor{black}{rgb}{0.0, 0.0, 0.0}
\definecolor{majorblack}{rgb}{0.0, 0.0, 0.0}

\IEEEraisesectionheading{
\section{Introduction}\label{sec:introduction}}
\IEEEPARstart{C}{urrent} deep neural networks have enabled great success in various computer vision tasks {\cite{ImagenetClassification_NeurIPS12,ResNet_CVPR16}}, based on the large-scale and computationally expensive training data {\cite{ImageNet_IJCV15,COCO_ECCV14,WePerson_MM21}}. However, in real world, data are usually distributed across numerous participants (\eg, mobile devices, organizations, cooperation). Due to the ever-rising privacy concerns and strict legislations {\cite{GDPR17}}, the traditional center training paradigm which requires to aggregate data together, is prohibitive. Driven by such realistic challenges, federated learning {\cite{FedAvg_AISTATS17,FederatedMLConApp_TIST19,LAQ_PAMI20,OFL_PAMI21}} has been popularly explored as it can train a global model for different participants rather than centralize any private data. Therefore, it provides an opportunity for privacy-friendly collaborative machine learning paradigm and has shown promising results to facilitate real world applications, such as keyboard prediction {\cite{FLKeyboard_arXiv18}}, medical analysis{\cite{FedDG_CVPR21}} and humor recognition {\cite{FLHumorReco_IJCAI2020}}.

\begin{figure}[t]
\begin{center}
\includegraphics[width=\linewidth]{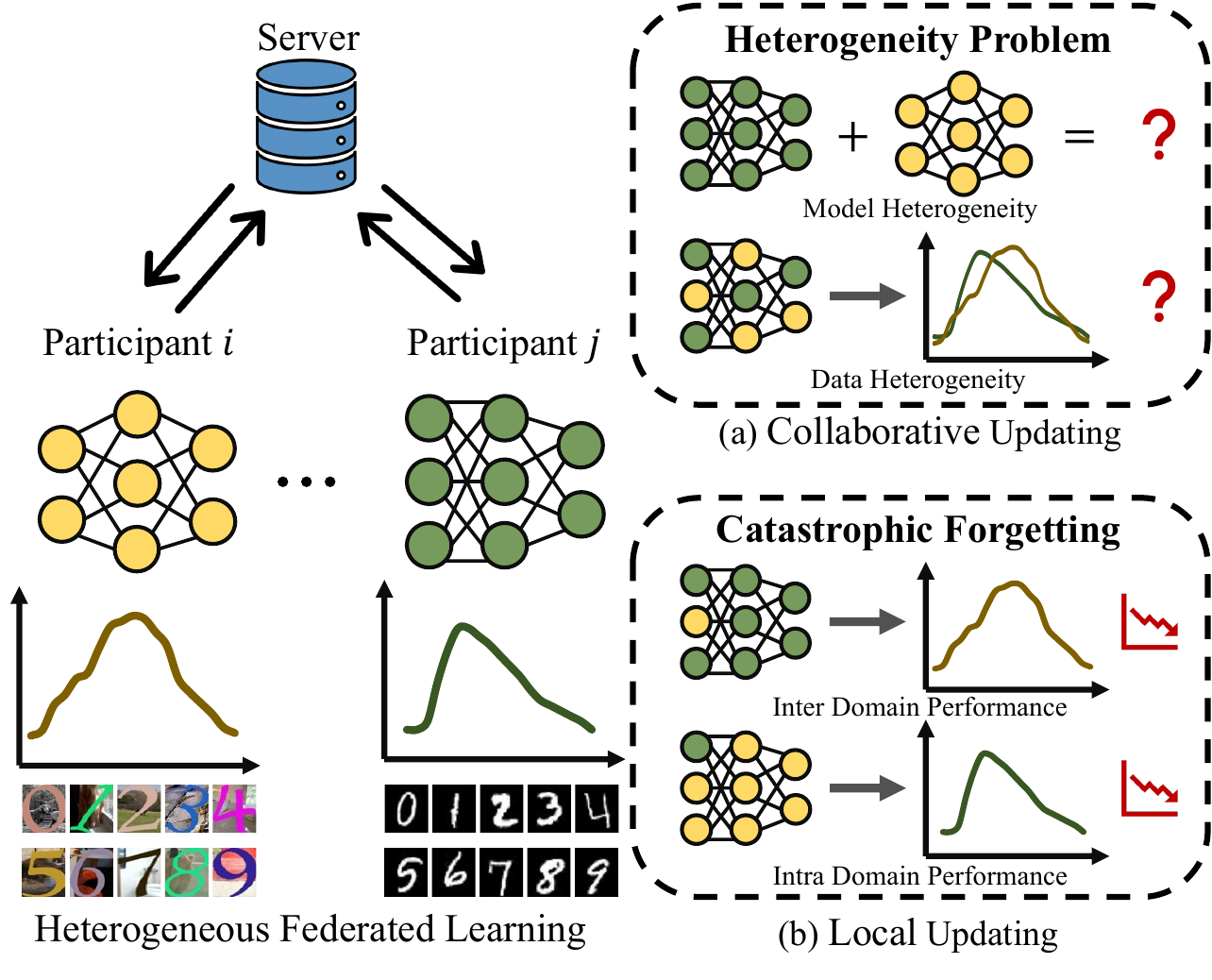}
\end{center}
\vspace{-20pt}
\caption{\textbf{Problem illustration of heterogeneous federated learning.} (a) In collaborative updating, how to handle the communication problem of heterogeneous models and learn a generalizable representation under heterogeneous data (domain shift)? (b) In local updating, how to effectively alleviate catastrophic forgetting to present stable and satisfactory performance in both inter- and intra-domains?}
\vspace{-20pt}
\label{fig:problem}
\end{figure}

Despite the great success afforded by federated learning, there remains fundamental research problems {\cite{Advances_arXiv19}}. One key challenge  is  \textbf{data heterogeneity} {\cite{FLONNonIID_TKDE22,HFL_CSUR23}}. In practical, private data distributions from different participants are inherently non-i.i.d (independently and identically distributed) since data are usually collected from different sources under varying scenarios. Existing works {\cite{2PersFL_TNNLS22}} have indicated that under heterogeneous data, standard federated learning methods such as FedAvg {\cite{FedAvg_AISTATS17}} inevitably faces performance degradation or even collapse. A resurgence of methods mainly focus on incorporating extra proximal terms to regularize the local optimization objective close to the global optima {\cite{FedProx_MLSys2020,FedCurv_NeurIPSW19,MOON_CVPR21}}. However, they ignore the fact that there exists \textbf{domain shift}, where the private data feature distribution of different participants may be strikingly distinct {\cite{DatasetShiftML_2009,SurveyonTF_TKDE09,UniViewonDatasetShift_PR12}}. Thus, the optimization objective is required to ensure both generalization performance on others domains and discriminability ability on itself domain. However, existing methods mostly focus on facilitate performance on intra-domain and would suffer from drastic performance drop on others domains with disparate feature distribution. As a result, forcing private model to acquire generalizable ability under domain shift is important and meaningful. 
The other inescapable and practical challenge is \textbf{model heterogeneity}. Specifically, participants have the  option of architecting private models rather than agreeing on the pre-defined architecture because they may have different design criteria, distinctive hardware capability {\cite{Fbnet_CVPR19}} or intellectual property rights {\cite{IPR_06}}. Thus, taking into account data heterogeneity and model heterogeneity together, previously aforementioned methods are no longer feasible because they are based on sharing consistent model architecture for parameter averaging operation. In order to address this problem, existing efforts can be taxonomized into two major categories: shared global model {\cite{FML_arXiv20,FLwiMoE_2020,LGFEDAVG_NeurIPS20}} and knowledge transfer {\cite{FedMD_NeurIPS19,Cronus_arXiv19,FedDF_NeurIPS20,FedGKT_NeurIPS20,RHFL_CVPR22}}. However, leveraging extra global model not only raises the communication cost but also necessitates additional model structure in participant side. For knowledge transfer methods, provided with unlabeled public data or artificially generated data, they utilize knowledge distillation {\cite{ModelComp_KDD06,KD_arXiv15}} to transfer the logits output on public data among participants. However, this strategy is subject to a significant limitation: they conduct knowledge transfer via logits output, which is in high semantic level {\cite{DKD_CVPR22}}. Directly aligning distribution probably provides confusing optimization direction {{\cite{DetectOOD_NeurIPS18,ConfinEsit_CVPR20}}} since the category of public data is unknowable and is  possible inconsistent with private data.
Thus, considering data and model heterogeneity together, an essential issue has long been overlooked: \textsl{(a) How to achieve a better generalizable representation in heterogeneous federated learning?} This problem is shown in \cref{fig:problem} (a).

{\color{black}Besides, in domain shift, data are typically from various domains. Thus, private models are required to maintain satisfactory performance  not only on the intra-domain but also on the inter-domain. However, a major hindrance is derived from the federated learning paradigm. In general, federated learning could be regarded as a cyclic process with two steps: collaborative updating and local updating {\cite{FedAvg_AISTATS17,FederatedMLConApp_TIST19}}. Specifically, during collaborative updating, participants learn from others and incorporate inter-domain knowledge and probably overlap intra-domain ability.
Then in local updating, the model is solely trained on private data. If optimized for too many epochs, it would be prone to overfit intra-domain knowledge and forgets previously learned inter-domain knowledge, leading to inter-domain catastrophic forgetting. In contrast, with too limited training epochs, it would perform poorly in the intra-domain, resulting in intra-domain catastrophic forgetting. Collaboratively considering the above two cases,  \textbf{catastrophic forgetting} {\cite{CataInter_PLM89,Connectionist_PR90,EmpInvCataFor_arXiv13}} presents  an important obstacle towards learning a generalization model in heterogeneous federated learning. }Related methods mainly focus on introducing parameter stiffness to regulate learning objective  {\cite{FedProx_MLSys2020,FedCurv_NeurIPSW19,pFedME_NeurIPS20,MOON_CVPR21,RHFL_CVPR22}}, which either rely on model homogeneity or do not take full advantage of the underlying inter-domain knowledge provided from models which are optimized in collaborative learning. Consequently, a natural question arises: \textsl{(b) How to effectively balance multiple domains knowledge to alleviate catastrophic forgetting?} An illustration of these two problems is presented in the \cref{fig:problem}.

To acquire generalization ability in heterogeneous federated learning, inspired by self-supervised learning with information maximization{\cite{WMSE_ICML21,Barlow_ICML21,vicreg_ICLR22}}, a preliminary version of this work published in CVPR 2022 \cite{FCCL_CVPR22} explores \textbf{\FCCM} (\FCCMab). In particular, we aim to make same dimensions of logits output highly correlated to learn class invariance and decorrelate pairs of different dimensions within logits output to stimulate the diversity of different classes. Moreover, representation  (embedding feature) exhibits generic information related to structural information {\cite{CRD_ICLR20}}, but is in lack of effective utilization. This motivates us to capture more significant information in other participants models representation. We propose \textbf{\FISL} (\FISLab), which leverages unlabeled public data to innovatively align instance similarity distribution among participants. Specifically, for each unlabeled public instance, private model calculates feature vector and obtains its similarity with respect to other instances in the same batch. We convert the individual feature distribution into similarity probability distribution. Then, we transfer that knowledge, so that participants are required to mimic the same similarity distribution. This approach offers a practical solution to realize feature level knowledge communication for differential feature under model heterogeneous federated learning. Besides, we believe that \FCCMab{} is essential because it optimizes classifier in collaborative learning to alleviate its bias towards local data distribution {\cite{CCVR_NeurIPS21,FedUFO_ICCV21}}. 
{\color{majorblack}
To this end, we develop both \FCCM{} and \FISL{} modules in heterogeneous federated learning, which handle the communication problem for heterogeneous models via achieving both logits and feature-level knowledge communication to acquire a more generalizable representation in heterogeneous federated learning.
}


To alleviate inter-domain knowledge forgetting in local updating, we propose a heterogeneous distillation to extract useful knowledge from respective previous model after communicating with other participants, which captures inter-domain knowledge. To maintain intra-domain discriminability, we intuitively leverage label supervision {\cite{CE_AOR05}}. However, there exists optimization conflict when simultaneously maintaining inter- and intra-domains knowledge.
Specifically, we observe that typical knowledge distillation {\cite{KD_arXiv15,BAN_ICML18}} can be viewed as  target distillation (ground-truth class prediction) and non-target distillation (residual classes prediction component).
We conjecture that non-target distillation depicts the classes relation provided by previous model which contains privileged inter-domain information {\cite{RNNwtKD_ICASSP16,DoesKDWork_NeurIPS21}}. 
For target distillation, it concentrates on intra-domain knowledge since it represents the specific prediction on ground-truth label. However, previous model could not consistently provide higher confidence on target class than current updating model, leading to contradictory gradient direction with respective label.  
In our conference version {\cite{FCCL_CVPR22}}, we leverage pretrained intra-domain model to alleviate this problem because it normally presents high confidence on target label, offering stable constraint.
In this journal version, we propose \textbf{\FNTD} (\FNTDab) to separately distill non-target information from previous model, which better preserves inter-domain knowledge and  ensures generalization ability while improving intra-domain discriminability. Compared with our conference version, \FNTDab{} not only avoids the requirement of a well-pretrained model but also improves the efficiency of inter-domain knowledge transfer in local updating. 

This paper builds upon on our conference paper \cite{FCCL_CVPR22} with extended contributions, which are listed in the following:

\begin{itemize} 
\item We introduce an effective heterogeneous feature knowledge transfer strategy in federated learning, \FISL{} (\FISLab{}), to build the connection among heterogeneous participants without additional shared network structure design. Through aligning the instance similarity distribution among unlabeled public data, heterogeneous models realize feature information communication. Collaborated with proposed \FCCM{} (\FCCMab{}), we realize both logits and feature levels knowledge communication, resulting in better generalizable ability.
\item We disentangle the typical knowledge distillation and develop \FNTD{} (\FNTDab{}) to better preserve inter-domain knowledge in the local updating stage. Cooperated with private data label supervision (intra-domain information) constraint, it effectively balances multiple domains knowledge and thus addresses  the catastrophic forgetting problem, improving both inter- and intra-domains performance. 
\item We provide a unified benchmark for both heterogeneous and homogeneous federated learning, supporting various state-of-the-art methods. We rigorously validate proposed method on massive scenarios (\ie, \digits{} {\cite{MNIST_IEEE98,USPS_PAMI94,svhn_NeurIPS11,syn_arXiv18}}, \officecaltech{} {\cite{OffCaltech_CVPR12}}, \officeTO{} {\cite{office31_ECCV10}} and \officehome{} {\cite{officehome_CVPR17}}) with unlabeled public data (\ie, \cifarhun{} \cite{cifar_Toronto09}, \tyimagenet{} \cite{ImageNet_IJCV15} and \fashionmnist{} \cite{fashionmnist_arXiv17}). We further provide detailed ablation studies and in-depth discussions to validate its effectiveness.
\end{itemize}

\section{Related Work}\label{sec:related}

\subsection{Heterogeneous Federated Learning}
\noindent \textbf{Data Heterogeneity}. A pioneering work proposed the currently most widely used algorithm, \fedavg {\cite{FedAvg_AISTATS17}}. But it suffers  performance drop on non-i.i.d data (data heterogeneity). Shortly after, a large body methods {\cite{FedProx_MLSys2020,FedCurv_NeurIPSW19,pFedME_NeurIPS20,MOON_CVPR21,RHFL_CVPR22,FSMAFL_ACMMM22,FGSSL_IJCAI23}} research on non-i.i.d data. These methods mainly focus on label skew with limited domain shift. However, when private data sampled from different  domains, these works do not consider inter-domain performance but only focus on learning an internal model. Latest researches have studied related problems on domain adaptation for target domain {\cite{FADA_ICLR20}} and domain generalization for unseen domains {\cite{FedDG_CVPR21}}. However, collecting enough data in  target domain can be time-consuming. Meanwhile, considering the performance on unknown domains is an idealistic setting. For more realistic settings, participants are probably more interested in the performance on others domains, which could directly improve economic benefits. In this paper, we focus on improving inter-domain performance under domain shift.

\noindent \textbf{Model Heterogeneity}.
With the demand for unique models, model heterogeneous federated learning has been an active research field. One Direction is introducing shared extra model {\cite{FML_arXiv20,LGFEDAVG_NeurIPS20}}. However, these techniques may not be applicable when considering  additional computing overhead and expensive communication cost. Recently, A line of work {\cite{FedMD_NeurIPS19,Cronus_arXiv19,CFD_arXiv20}} operates on labeled public data (with similar distribution) via knowledge distillation {\cite{ModelComp_KDD06,KD_arXiv15}}. Therefore, these approaches heavily rely on the quality of labeled public data, which may not always be available on the server. Latest works \cite{FedDF_NeurIPS20,FedKT_arXiv20,FEDGEN_ICML21} have proven the feasibility to do distillation on unlabeled public data or synthetic data. However, these methods reach semantic information consistency on unlabeled public data, which are not suitable to learn a generalizable representation and thus lead to a sub-optimal inter-domain performance. In this paper, based on unlabeled public data, we construct cross-correlation matrix and measure instance similarity on unlabeled public data to reach generalizable ability.

\subsection{Catastrophic Forgetting}
Catastrophic forgetting {\cite{Connectionist_PR90,EmpInvCataFor_arXiv13}} has been an essential problem in incremental learning {\cite{silver2002task,MAS_eccv18}} when models continuously learn from a data stream, with the goal of gradually extending acquired knowledge and using it for future {\cite{CataInter_PLM89,EmpInvCataFor_arXiv13}}.  Existing incremental learning works can be broadly divided into three branches {\cite{AConLearSurvey_TPAMI21}}: replay methods {\cite{iCARL_CVPR17,Dark_NeurIPS20}}, regularization-based methods {\cite{LwF_TPAMI17,PASS_CVPR21}} and parameter isolation methods {\cite{PNN_arXiv16,DER_CVPR21}}. As for federated learning, data are distributed rather than sequential like incremental learning. Despite these differences, both incremental learning and federated learning share a common challenge - how to balance the knowledge from the different data distributions.  
{\color{black} Unlike incremental learning methods, we focus on alleviating catastrophic forgetting in distributed data rather than time series data. In particular, we expect to alleviate both inter- and intra-domains catastrophic forgetting in the local updating stage to acquire a generalization model.}

\subsection{Self-Supervised Learning}
Self-supervised learning has emerged as a powerful method for learning useful representation without supervision from labels, largely reducing the performance gap between supervised models on various downstream vision tasks. Many related methods are based on contrastive methods {\cite{instdisc_cvpr18,InvaSpread_CVPR19,AIISMang_PAMI20,MOCO_CVPR20,SimCLR_ICML20}}. This kind method contrasts positive pairs against negative pairs, where positive pairs are often formed by same samples with different data augmentations and negative pairs are normally other different samples. Recently, another line of works {\cite{BYOL_NeurIPS20,SimSiam_CVPR21}} employs asymmetry of the learning update (stop-gradient operation) to avoid trivial solutions. Besides, a principle to prevent collapse is to maximize the information content of the embeddings {\cite{WMSE_ICML21,Barlow_ICML21,vicreg_ICLR22}}.
The key difference between \FCCM{} (\FCCMab{}) and aforementioned efforts  is that ours is designed for federated setting rather centralized setting. Inspired by self-supervised learning, We construct the comparison among heterogeneous models in federated learning.

\subsection{Knowledge Distillation}
Knowledge distillation aims to transfer knowledge from one network to the other  {\cite{ModelComp_KDD06,KD_arXiv15}}. The knowledge from the teacher network can be extracted and transferred in various ways. Related approaches can be mainly divided into three directions: logits distillation {\cite{KD_arXiv15,PSKD_ICCV21,KR_AAAI21,DKD_CVPR22}}, feature distillation {\cite{Fitnets_ICLR15,SemCKD_AAAI21}} and relation distillation {\cite{GiftKD_CVPR17,RKD_CVPR19}}. In this work, we propose \FISL{} (\FISLab{}) to implement multi-party rather than pair-wise heterogeneous feature communication in collaborative updating. Furthermore, we investigate how to better maintain inter-domain knowledge in local updating and develop \FNTD{} (\FNTDab{}) to disentangle logits distillation, which assures inter-domain performance without the heavy computational strain.

\begin{figure*}[t]
\centering
\includegraphics[width=0.95\linewidth]{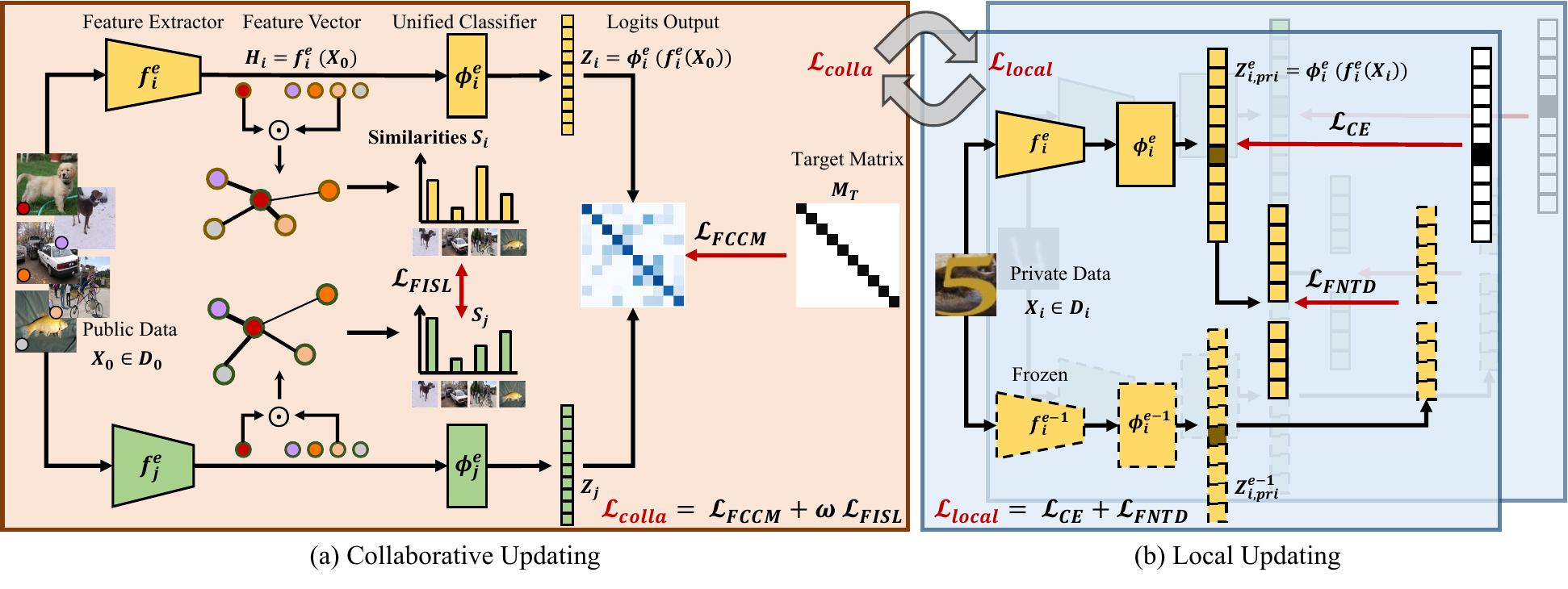}
\put(-270,97){\scriptsize{\cref{eq:fccm_loss}}}
\put(-337,139){\scriptsize{\cref{eq:simmat_new}}}
\put(-343,93){\scriptsize{\cref{eq:fisl}}}
\put(-237,153){\scriptsize{\cref{eq:col_loss}}}
\put(-180,153){\scriptsize{\cref{eq:localobj}}}
\put(-84,80){\scriptsize{\cref{eq:fntd}}}
\put(-80,129){\scriptsize{\cref{eq:localCE}}}
\vspace{-15pt}
\caption{\textbf{Schematization of \oursabbrv{}}. Our method solves heterogeneity problem and catastrophic forgetting in federated learning.
(a) Collaborative updating: we propose \FCCM{} (\FCCMab{} \cref{sec:3.1}) via constructing cross-correlation matrix $\mathcal{M}_i$ to target matrix, $\mathcal{M}_T = 2 \times eye(C)-ones(C)$, where on-diagonal is $1$, off-diagonal is $-1$. Besides, we introduce \FISL{} (\FISLab{} \cref{sec:3.2}) to realize heterogeneous feature communication.
(b) Local updating: we decouple typical knowledge distillation and develop \FNTD{} (\FNTDab{} \cref{sec:3.3}) to better preserve inter-domain knowledge. Moreover, we leverage private data label signal to maintain intra-domain performance. Best viewed in color. Zoom in for details.}  
\label{fig:framework}
\vspace{-15pt}
\end{figure*}

\section{Methodology}\label{sec:Methodology}
\textbf{Problem Statement and Notations.}
Following the standard federated learning setup, there are $K$ participants (indexed by $i$). The $i^{th}$ participant has a local model $\theta_i$ and private data $D_i =\{(X_i,Y_i)|X_i \in \mathbb{R}^{N_i \times O}, Y_i \in \mathbb{R}^{N_i \times C}\}$, where $N_i$ denotes the scale of private data, $O$ represents input size and $C$ is defined as classification categories. {\color{majorblack}We further consider local model as two components: feature extractor and unified classifier. The feature extractor $f: \mathcal{X} \rightarrow \mathcal{H}$, maps input image $x$ into a compact $d$ dimensional feature vector $H = f(x) \in \mathbb{R}^d$. A unified classifier $\phi:  \mathcal{H} \rightarrow  \mathcal{Z}$, produces probability output, $Z=\phi(f(x))\in \mathbb{R}^{C}$ as prediction for $x$.  Thus, the overall private model is denoted as $\theta_i=(f_i,\phi_i)$.} In heterogeneous federated learning, \textbf{data heterogeneity} and \textbf{model heterogeneity} are defined as:

\begin{itemize}
\item \textbf{Data heterogeneity}: $P_i(X|Y)\neq P_j(X|Y)$. There exists domain shift among private data, \ie, conditional distribution $P(X|Y)$ of private data vary across participants even if $P(Y)$ is shared. Specifically, the same label $Y$ has distinctive feature $X$ in different domains. 

\item \textbf{Model heterogeneity}: $Shape(\theta_i) \neq Shape(\theta_j)$. Participants architect models independently, \ie, the feature extractor $f$ are different such as \resnet {\cite{ResNet_CVPR16}, \efficientnet {\cite{EfficientNet_ICML19}}} and \mobilenet {\cite{MobileNet_arXiv17}}, and  the corresponding classifier module is also differential.
\end{itemize}
The unlabeled public dataset $D_0 = \{X_0|X_0 \in \mathbb{R}^{N_0 \times O}\}$ is adopted to achieve heterogeneous models  communication, which is relatively easy to access in real scenarios, \eg, existing datasets {\cite{COCO_ECCV14,ImageNet_IJCV15,WePerson_MM21}} and web images{\cite{Webvision_arXiv17}}. 
{\color{majorblack}
The goal for $i^{th}$ participant is to reach communication and learn a model $\theta_i$ with generalizable ability. Besides, with \textbf{catastrophic forgetting}, the optimization objective is to present both higher and stabler inter- and intra-domains performance. We provide the notation table in \cref{tab:notation} for better understanding.
}

\begin{table}[t]
\centering
\caption{
{\color{majorblack}
\small{
\textbf{Notations} table.
}
}
}
\vspace{-10pt}
\label{tab:notation}
\resizebox{\linewidth}{!}{
		\setlength\tabcolsep{2pt}
		\renewcommand\arraystretch{1.2}
\begin{tabular}{clIcl}
\hline\thickhline
\rowcolor{lightgray}
 & Description &  & Description\\ \hline\hline
$K$ & Participant scale&
$i$ & Participant index 
\\
$D_i$ & Private data set of $i$ client &
$B_0$ & Private data $D_i$ batch 
\\
$C$ & Classification categories &
$O$ & Input size
\\
$\theta$ & Neural network &
$f$ & Feature extractor 
\\
$\phi$ & Unified classifier &
$H$ & Feature embedding 
\\
$Z$ & Logits output &
$d$ & Feature dimension 
\\ 
$N_i$ & Scale of $D_i$ &
$D_0$ & Unlabeled public data
\\
$B_0$ & Public data $D_0$ batch & 
$M_i$ &  Cross-correlation matrix (\cref{eq:avgcrosscorr}) 
\\
$S_i$ & Instance similarity distribution(\cref{eq:simmat_old}) &
$\mathbf{p}$ & Predictive distribution (\cref{eq:kd_ori_form}) 
\\
$E$ & Communication epochs & 
$T$ & Local rounds 
\\
\hline
\end{tabular}%
}
\vspace{-15pt}
\end{table}

\subsection{\FCCM}\label{sec:3.1}
\textbf{Motivation of Dimension-Level Operation}.
Motivated by {information maximization} {\cite{Infobot_arXiv2000,Barlow_ICML21,vicreg_ICLR22}}, a generalizable representation should be as informative as possible about image, while being as invariant as possible to the specific domain distortions that are applied to this sample. In our work, domain shift leads distinctive feature $X$ for the same label $Y$ in different domains. 
Therefore, the distribution of logits output along the batch dimension in different domains is not identical.
Moreover, different dimensions of logits output are corresponding to distinct classes. 
Thus, we encourage the invariance of the same dimensions and the diversity of different dimensions.
Private data carries specific domain information and is under privacy protection, which is not suitable and feasible to conduct collaborative updating. Therefore, we leverage unlabeled public data, which are normally generated and collected from multi domains. Hence, private models are required to reach logits output invariant to domain distortion and decorrelate different dimensions of logits output on unlabeled public data.

\noindent \textbf{Construct Federated Cross-Correlation Matrix}.
Specifically, we obtain the logits output from $i^{th}$ participant: $Z_i=\phi (f(X_0)) \in \mathbb{R}^{|B_0| \times C}$ on unlabeled public data $D_0 (X_0)$ with batch  $B_0$. For $i^{th}$ and $j^{th}$ participant, the logits output is $Z_i$ and $Z_j$, respectively. 
Notably, considering the computing burden on the server side, we calculate average logits output: $\overline{Z}=\frac{1}{K}\sum_iZ_i$. Then, we calculate  cross-correlation matrix, $\mathcal{M}_i$ for $i^{th}$ participant with average logits output as:
{\color{majorblack}\begin{equation}\small
\setlength\abovedisplayskip{0pt} \setlength\belowdisplayskip{0pt}
\mathcal{M}_i^{uv} \triangleq \frac{
\sum_b (\Phi(Z_i)^{b,u} ~~ \Phi(\overline{Z})^{b,v})}
{\sqrt{\sum_b {\Phi(Z_i)^{b,u}}^2} \sqrt{\sum_b {\Phi(\overline{Z})^{b,v}}^2}}.
\label{eq:avgcrosscorr}
\end{equation}}
{\color{majorblack}The $b$ indexes batch samples,  $u,v$ index the  dimension of logits output and $\Phi$ is the normalization operation along the batch dimension with mean subtraction. $\mathcal{M}_i$ is a square matrix with size of output dimensionality, $C$ and is comprised between -1 (\ie, dissimilarity) and 1 (\ie, similarity).}
Then, the \FCCMab{} loss for $i^{th}$ participant is defined as:
\begin{equation}\small
\setlength\abovedisplayskip{0pt} \setlength\belowdisplayskip{0pt}
\mathcal{L}_{{\FCCMab}} =  {\sum_u  (1-\mathcal{M}_i^{uu})^2} + \lambda {\sum_{u}\sum_{v \neq u} {(1+\mathcal{M}_i^{uv}})^2},
\label{eq:fccm_loss}
\end{equation}
where $\lambda$ is a positive constant trading off the importance of the first and second terms of  loss. Naturally, when on-diagonal terms of the  cross-correlation matrix take the value $+1$, it encourages the logits output from different participants to be similar; when off-diagonal terms of the  cross-correlation matrix take value $-1$, it encourages the diversity of logits output since different dimensions of these logits output will be uncorrelated to each other.

\begin{figure*}[t]
	\begin{center}
		\includegraphics[width=0.95\linewidth]{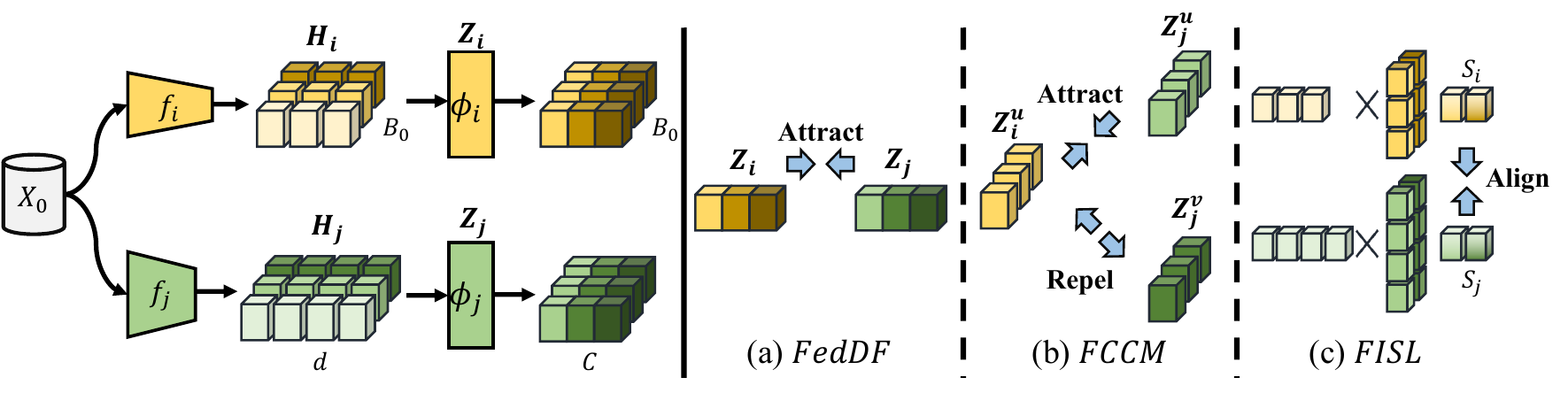}
		\put(-214,10){\small{\cite{FedDF_NeurIPS20}}}
		\put(-127,10){\small{\cite{FCCL_CVPR22}}}
		\put(-24,99){\scriptsize{\cref{eq:simmat_new}}}
	\end{center}
    \vspace{-20pt}
    \caption{\color{black}\textbf{Conceptual Illustration.} The unlabeled public data $D_0(X_0)$ with batch size $B_0$ are fed into different models. 
    The feature vector $H_i = f_i(X_0)$ has $d$ dimensions, which is inconsistent for different models.
    The logits output $Z_i=\phi_i(f_i(X_0))$ has $C$ dimensions, which is same due to the consistent classification categories. 
    (a) \feddf {\cite{FedDF_NeurIPS20}} calculates the distribution divergence where instance-wise normalized \textbf{logits output} is compared inside a batch. 
    (b) \FCCMab{} {\cite{FCCL_CVPR22}} learns invariance in same dimensions and decorrelates pairs of different dimensions on the batch-wise normalized \textbf{logits output} in \cref{eq:avgcrosscorr}. Thus, for each local logits dimensional output, it aims to attract with same dimension  and repel different dimensions with other clients
    See \cref{sec:3.1} for details. 
    (c) \FISLab{} further leverages \textbf{feature vector} to calculate instance similarity distribution $S_i$ by \cref{eq:simmat_old} and \cref{eq:simmat_new}. Then, optimize present model to align it with others via \cref{eq:fisl}. See \cref{sec:3.2} for details. Best viewed in color.}
	\label{fig:ConceptIllu}
    \vspace{-10pt}
\end{figure*}

\noindent \textbf{Comparison with Analogous Methods}. 
\fedmd{} \cite{FedMD_NeurIPS19} minimizes mean square error on relative public data with annotation. \feddf{} \cite{FedDF_NeurIPS20} reaches logits output distribution consistency on unlabeled public data.  However, \FCCMab{} expects to achieve correlation of same dimensions but decorrelation of different dimensions on unlabeled public data. Besides, we operate along the batch dimension, which means that we view unlabeled public data as ensemble rather than individual sample. It is advantageous to eliminate anomalous sample disturbance. We further illustrate the conceptual comparison between  \feddf{} and \FCCMab{} in \cref{fig:ConceptIllu}.

\begin{algorithm}[t]
\SetNoFillComment
\SetArgSty{textnormal}
\small{\KwIn{Communication epochs $E$, local rounds $T$, participants number $K$, unlabeled public data $D_0(X_0)$, $i^{th}$ private data $D_i(X_i,Y_i)$ , hyper-parameter $\lambda, \mu, \omega,\tau$}}
\BlankLine

\For {$e=1, 2, ..., E$}{
    
    \For {$i=1, 2, ..., K$}{
    $H_i = f_i^e(X_0)$  \tcp*{feature vector} 
    $Z_i = \phi_i^e(H_i)$ \tcp*{logits output}
    $S_i \leftarrow (H_i,\mu)$ by \cref{eq:simmat_old} and \cref{eq:simmat_new} \tcp*{instance similarity }
    }
    
    $\overline{S} = \frac{1}{K}\sum_iS_i$ \
    $\overline{Z} = \frac{1}{K}\sum_iZ_i$
    
    \For {$i=1, 2, ..., K$ \textbf{in parallel}}{
    
    $\mathcal{L}_{{\FCCMab{}}} \leftarrow$ \textbf{{\FCCM{}}}~($Z_i$, $\overline{Z}$)  
    
    $\mathcal{L}_{{\FISLab{}}} \leftarrow$ \textbf{{\FISL{}}}~($S_i$, $\overline{S}$)
    
    $\mathcal{L}_{colla} =\mathcal{L}_{{\FCCMab{}}} + \omega \mathcal{L}_{{\FISLab{}}}$ (\cref{eq:col_loss}) 
    
    $\theta_i^e \leftarrow \theta_i^e - \eta \nabla \mathcal{L}_{colla}$
    
    $\theta_i^{e+1} \leftarrow $ \textbf{\FNTD{}} ($\theta_i^{e-1}$, $D_i$)
    }
}
return $\theta_i^{E+1}$

\BlankLine
\textbf{\FCCM}~($Z_i$, $\overline{Z}$)  

{\color{black}{See details in \cref{sec:3.1}}} 

    $\mathcal{M}_i \leftarrow (Z_i,\overline{Z})$ { by \cref{eq:avgcrosscorr}}
    
     $\mathcal{L}_{\FCCMab{}} \leftarrow (\mathcal{M}_i,\lambda)$ through \cref{eq:fccm_loss} 
     
return $\mathcal{L}_{\FCCMab{}}$

\BlankLine
\textbf{\FISL{}}~($S_i$, $\overline{S}$)  

{\color{black}{See details in \cref{sec:3.2}}} 

     $\mathcal{L}_{\FISLab{}} \leftarrow (S_i, \overline{S})$ via \cref{eq:fisl} 
     
return $\mathcal{L}_{\FISLab{}}$

\BlankLine

\textbf{\FNTD}~($\theta_i^{e-1}$, $D_i$): 

{\color{black}{See details in \cref{sec:3.3}}}

\For{$t = 1, 2, ..., T$}{

        $Z_{i,pri}^{e}=\phi_i^e(f_i^{e}(X_i))$

        $\mathcal{L}_{CE} \leftarrow \texttt{CE}(Z_{i}^e,Y_i)$ in \cref{eq:localCE}
        
        $Z_{i,pri}^{e-1}=\phi_i^{e-1}(f_i^{e-1}(X_i))$

        $\mathcal{L}_{\FNTDab{}} \leftarrow (Z_{i,pri}^{e},Z_{i,pri}^{e-1}, \tau)$ by \cref{eq:fntd}
        
        $\mathcal{L}_{local} = \mathcal{L}_{CE}+\mathcal{L}_{\FNTDab{}}$ (\cref{eq:localobj})
        
        $\theta_i^{e} \leftarrow \theta_i^{e}-\eta \nabla \mathcal{L}_{local}$
        
}

$\theta_i^{e+1} \leftarrow \theta_i^{e}$

return $\theta_i^{e+1}$ 
\caption{\oursabbrv{}}
\label{alg:ours}
\end{algorithm}

\subsection{\FISL}\label{sec:3.2}
\noindent \textbf{Feature-Level Communication Dilemma}.
It is known that embedding feature can be convenient to express
many generic priors about the world {\cite{RepreSurvey_PAMI13,CRD_ICLR20}}, \ie, priors that are not task-specific but would be likely to be useful for generalization ability. Intuitively, it is beneficial to conduct feature-level communication in federated learning. However, in model heterogeneous federated learning, embedding features are in distinctive structure for different participants. 

\noindent \textbf{Instance-wise Similarity}.
Due to the heterogeneous feature, we formulate a novel and simple approach for the feature-wise communication on the basis of instance similarity distribution on unlabeled public data. Formally, given  $B_0$ a batch of unlabeled public data $D_0$, the $i^{th}$ private model $\theta_i$  maps  them into the embedding feature vector $H_i = f(B_0) \in \mathbb{R}^{|B_0|\times d}$. Let $S_i$ represents the similarity distribution of feature vector among unlabeled public data batch computed by the $i^{th}$ private model and is defined as:
{\color{majorblack}
\begin{equation}\small
\setlength\abovedisplayskip{0pt} \setlength\belowdisplayskip{0pt}
\begin{split}
S_i \triangleq \frac{H_i \cdot H_i^T}{\mu ||H_i||_2 ||H_i||_2} \in \mathbb{R}^{|B_0|\times |B_0|},
\label{eq:simmat_old}
\end{split}
\end{equation}}
where $|| \cdot ||_2$ represents $\ell_2$ norm, $\mu$ is the soften hyper-parameter, $(\ )^T$ denotes transpose operation and $(\ \cdot \ )$ means inner product. Note that, we eliminate the diagonal value in $S_i$, which is the similarity value with itself and poses large value, resulting in dominant effect for the whole similarity distribution. Hence, the $S_i$ is rewritten into:
{\color{majorblack}
\begin{equation}\small
\setlength\abovedisplayskip{0pt} \setlength\belowdisplayskip{0pt}
\begin{split}
S_i & = S_i \odot \left[ \begin{array}{cccc}
0 & 1 & \cdots& 1 \\
1 & 0 & \cdots  & 1 \\
\vdots & \vdots & \ddots & 1 \\
1 & 1 & 1  & 0 
\end{array} 
\right ] \\
S_i & = S_i[S_i \neq 0] \in \mathbb{R}^{(|B_0|-1) \times (|B_0|-1)}.
\label{eq:simmat_new}
\end{split}
\end{equation}}
Similarly, it is important to control the server-side computing overhead, we measure the average similarity distribution: $\overline{S}=\frac{1}{K}\sum_iS_i$.  Finally, the \FISL{} (\FISLab{}) optimization objective  for $i^{th}$ participant is:
{\color{majorblack}
\begin{equation}\small 
\setlength\abovedisplayskip{0pt} \setlength\belowdisplayskip{0pt}
\mathcal{L}_{\FISLab{}}=\sigma(\overline{S},dim\!=\!1)\log \frac{\sigma(\overline{S},dim\!=\!1)}{\sigma(S_i,dim\!=\!1)},
\label{eq:fisl}
\end{equation}}
where $\sigma$ denote \texttt{softmax} function. As shown in \cref{eq:fisl}, We achieve the purpose of feature-level knowledge communication in model heterogeneous federated learning and lean more versatile information from others, bringing better generalizable ability. We further illustrate \FISLab{} in \cref{fig:ConceptIllu}.

\noindent \textbf{Overall Objective in Collaborative Updating}.
By combining $\mathcal{L}_{\FCCMab{}}$ in \cref{eq:fccm_loss} and $\mathcal{L}_{\FISLab{}}$ in \cref{eq:fisl}, the overall training target in collaborative updating becomes:
\begin{equation}\small 
\setlength\abovedisplayskip{0pt} \setlength\belowdisplayskip{0pt}
\mathcal{L}_{colla} = \mathcal{L}_{\FCCMab{}}+\omega  \mathcal{L}_{\FISLab{}}.
\label{eq:col_loss}
\end{equation}
\subsection{\FNTD}
\label{sec:3.3}
\textbf{Typical Knowledge Distillation}.
Leveraging the previous model optimized in collaborative learning to conduct knowledge distillation on present updating model is beneficial to alleviate inter-domain knowledge forgetting in local updating. Existing knowledge distillation mainly follow three lines: feature distillation, relation distillation and logits distillation {\cite{KDSurvey_IJCV_21}}. However, compared with collaborative updating phase where major computational cost is allocated to server rather participants, in local updating phase, private models are required to train for multiple epochs and the first two distillation methods pose markedly larger computational cost than logits distillation. Thus, leveraging logits knowledge distillation is feasible and practical in local updating. We denote $\mathcal{T}$ and $\mathcal{S}$  as  teacher and  student models. The logits knowledge distillation loss is defined as:
{\color{black}\begin{equation}\small
\setlength\abovedisplayskip{0pt} \setlength\belowdisplayskip{0pt}
    \begin{split}
    \mathcal{L}_{{KD}}&= \text{KL}(\mathbf{p}_{\mathcal{T}}|| \mathbf{p}_{\mathcal{S}}) \\
    {\mathbf{p}(x,\theta)} &=[{p}^{1}, .., {p}^{t}, .., {p}^{C}] \in \mathbb{R}^{1\times C},
    \end{split}
    \label{eq:kd_ori_form}
\end{equation}}
{\color{black}\begin{equation}\small
\setlength\abovedisplayskip{0pt} \setlength\belowdisplayskip{0pt}
\begin{split}
    \mathcal{L}_{{KD}} &=  \sum_{u}^{C} p_{\mathcal{T}}^{u,\tau}\log(\frac{p_{\mathcal{T}}^{u,\tau}}{p_{\mathcal{S}}^{u,\tau}})\\
    p^{u,\tau}(x,\theta) &= \frac{\exp(z^{u} / \tau)}{\sum_{v}^{C} \exp(z^{v} / \tau)}, z = \phi (f(x)),
\end{split}
    \label{eq:kd_component}
\end{equation}}
where  $z^u$ represents the logits output of $u^{th}$ class and $\tau$ is temperature hyper-parameter. Besides, in local updating period, leveraging private data label to construct CrossEntropy {\cite{CE_AOR05}} loss is helpful to overcome intra-domain knowledge forgetting {\cite{FedBN_ICLR21,FedMD_NeurIPS19,FedDF_NeurIPS20,FedDF_NeurIPS20}} and constructed as: 
\begin{equation}\small 
\setlength\abovedisplayskip{0pt} \setlength\belowdisplayskip{0pt}
\mathcal{L}_{CE}=-\bm{1}_{y}\log(\mathbf{p})=-\sum_{u}^C y^u\log(p^u).
\label{eq:localCE}
\end{equation}
We carry out the following optimization objective for the  $k^{th}$ participant in the local updating process:
{\color{black}\begin{equation}\small
\setlength\abovedisplayskip{0pt} \setlength\belowdisplayskip{0pt}
    \mathcal{L}_{{local}} = \mathcal{L}_{{CE}}+
    \tau^2  \mathcal{L}_{{KD}}(\mathcal{T}:\theta^{e-1}).
    \label{eq:orilocalobj}
\end{equation}}
The teacher model $\mathcal{T}$ is the previous model  $\theta^{e-1}$ after local updating from last communication epoch. However, for such training objective, there exists the optimization conflict. Specifically, We disentangle the logits distillation into target distillation (TD) and non-target distillation (NTD) as: 
\begin{equation}\small
\setlength\abovedisplayskip{0pt} \setlength\belowdisplayskip{0pt}
    \begin{split}
    \mathcal{L}_{{KD}}
    &= \underbrace{p_{\mathcal{T}}^{t}\log(\frac{p_{\mathcal{T}}^{t}}{p_{\mathcal{S}}^{t}})}_{\mathcal{L}_{{TD}}}+ \underbrace{\sum_{u \neq t}^{C} p_{\mathcal{T}}^{u}\log(\frac{p_{\mathcal{T}}^{u}}{p_{\mathcal{S}}^{u}})}_{\mathcal{L}_{{NTD}}}.
    \end{split}
    \label{eq:kd_decouple_form}
\end{equation}
{\color{black} As reflected in \cref{eq:kd_decouple_form}, target distillation provides the confidence degree on the ground-truth class and non-target distillation depicts the class relation, also reported in \cite{DKD_CVPR22,NKD_arXiv22}. But they are based on the traditional knowledge distillation setting and claim that these two parts are highly coupled. They decouple these two terms to better transfer knowledge from the teacher to the student. However, in our paper, we focus on the catastrophic forgetting problem in the local updating stage of federated learning and argue that the confidence on the target class $t$ is uncertain and would affect the backpropagation for the gradient as follows:}
\begin{equation}\small
\setlength\abovedisplayskip{0pt} \setlength\belowdisplayskip{0pt}
	\begin{split}
    \frac{\partial \mathcal{L}_{local}}{\partial z^{t}_\mathcal{S}} 
    &= \frac{\partial \mathcal{L}_{CE}}{\partial z^{t}_\mathcal{S}} + \frac{\partial{} \tau^2 \mathcal{L}_{KD}}{\partial z^{t}_\mathcal{S}} \\
    &= \tcbhighmath{\frac{\partial \mathcal{L}_{\text{CE}}}{\partial z^{t}_\mathcal{S}}} + \tcbhighmath[colback=green!8]{\frac{\partial{} \tau^2 \mathcal{L}_{\text{TD}}}{\partial z^{t}_\mathcal{S}}} + {\frac{\partial{} \tau^2 \mathcal{L}_{\text{NTD}}}{\partial z^{t}_\mathcal{S}}} \\
    &= \underbrace{\tcbhighmath{\frac{\exp(z^{u})}{\sum_{v}^{C} \exp(z^{v})}-1}}_{\in (-1,0)} + 
    \underbrace{\tcbhighmath[colback=green!8]{\tau(p_\mathcal{S}^t - p_\mathcal{T}^{t})}}_{\in (-\tau,\tau)} + 0.\\
    \label{eq:grad_conflict}
        \end{split}
\end{equation}
{\color{majorblack} $z^{t}_\mathcal{S}$ denotes the target class, $t$ logits output from the student model $\mathcal{S}$.
Notably, $p_\mathcal{T}^{t}$, the target class prediction from the teacher model is uncontrollable and probably leads to contradictory gradient direction  with the corresponding label, which hinders knowledge transfer. In particular, when the teacher model prediction on target class $t$ shows less confidence than the student model, it poses the inconsistent objective with the one-hot encoded label. This incongruity impedes fitting the intra-domain knowledge and decreases intra-domain confidence. Conversely, when teacher model prediction on target class $t$ exhibits higher confidence than the student model, it would increase the confidence degree on the intra-domain and thus hinder maintaining inter-domain knowledge. Hence, the student model struggles with the balance of under- and over-confidence problems.}

\noindent \textbf{\FNTD{}}. In this work, we develop \FNTD{} (\FNTDab) that addresses contradictory optimization objective in local updating phase. We disentangle knowledge distillation to fully transfer inter-domain information and hence remove potential negative  performance impact. The \FNTD{} objective for $i^{th}$ participant is as: 
\begin{equation}\small
\setlength\abovedisplayskip{0pt} \setlength\belowdisplayskip{0pt}
	\begin{split}
    \mathcal{L}_{\FNTDab{}} 
    &= \cancelto{0}{p_{\mathcal{T}}^{t}\log(\frac{p_{\mathcal{T}}^{t}}{p_{\mathcal{S}}^{t}})}+ \sum_{u \neq t}^{C} p_{\mathcal{T}}^{u}\log(\frac{p_{\mathcal{T}}^{u}}{p_{\mathcal{S}}^{u}})
    \\
    & =  \sum^{C}_{\color{red}{u \neq t}} p_{\mathcal{T}}^{u}\log(\frac{p_{\mathcal{T}}^{u}}{p_{\mathcal{S}}^{u}}).
    \label{eq:fntd}
    \end{split}
\end{equation}
$\mathcal{T}$ is the model after previous communication epoch ($\theta^{e-1}$), which provides inter-domain knowledge to handle inter-domain knowledge catastrophic forgetting  in local updating. We provide the detailed comparison between \FNTDab{} and typical KD in \cref{fig:ab_fntd}. Although $\theta^e$ after collaborative updating would involve more inter-domain knowledge, $\theta^{e-1}$ is optimized in local updating and provides customized inter-domain knowledge. We further empirically validate the superiority of $\theta^{e-1}$ in \cref{fig:teacherinFNTD}. Besides, we 
utilize the CrossEntropy {\cite{CE_AOR05}} to provide  intra-domain information. Finally, the local updating training target is formulated as:
\begin{equation}\small
\setlength\abovedisplayskip{0pt} \setlength\belowdisplayskip{0pt}
    \mathcal{L}_{{local}} = \mathcal{L}_{{CE}} + \mathcal{L}_{\FNTDab{}}(\mathcal{T}:\theta^{e-1}).
    \label{eq:localobj}
    \end{equation}

\begin{figure}[t]
\begin{center}
\includegraphics[width=\linewidth]{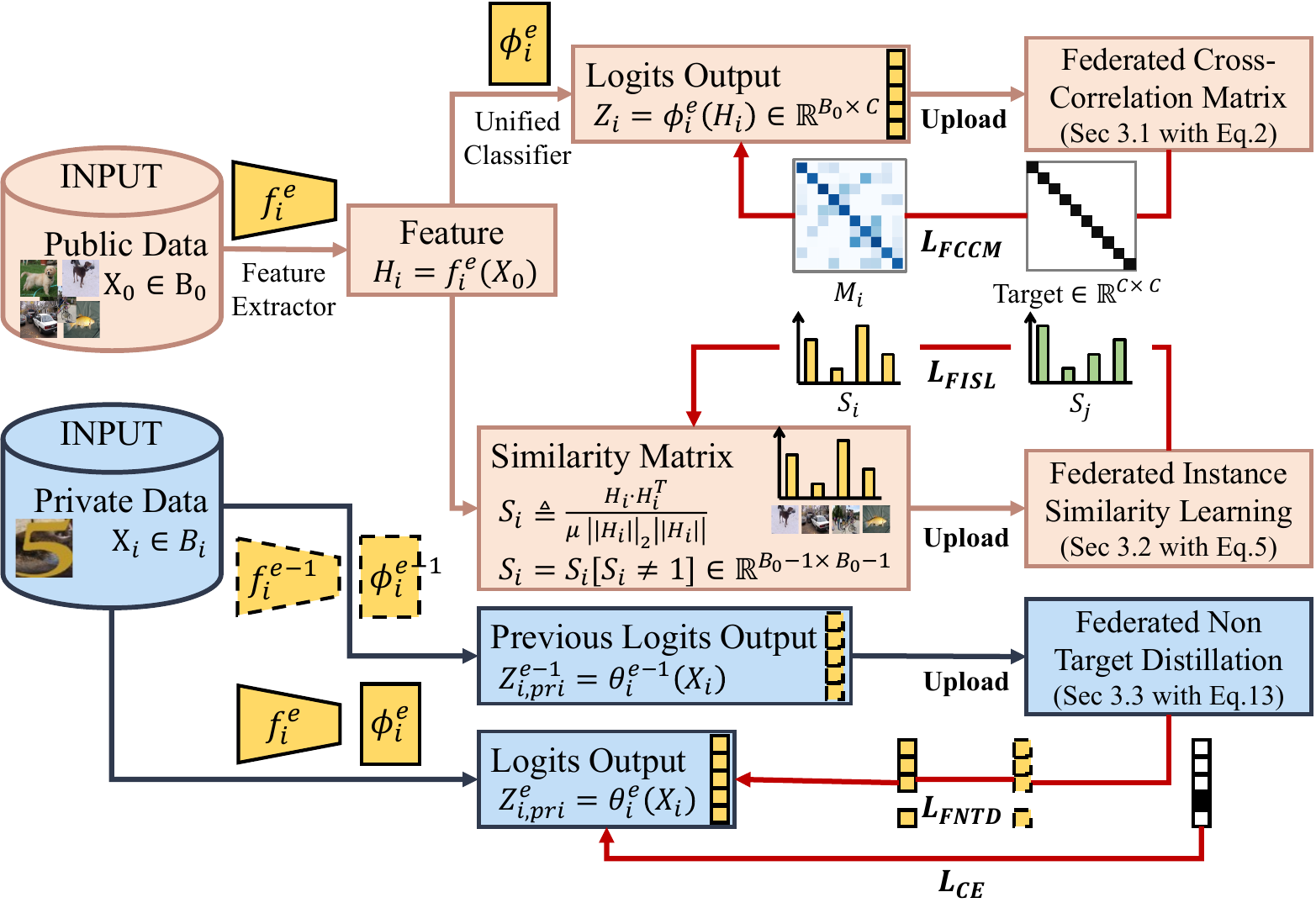}
\end{center}
\vspace{-15pt}
\caption{
{\color{majorblack}{
\textbf{The flow chart of our method}. Collaborative updating involves both Federated Cross-Correlation Matrix \cref{sec:3.1} and Federated Instance Similarity Learning \cref{sec:3.2} to achieve both feature and logits-level model heterogeneous communication. As for local updating, the Federated Non Target Distillation \cref{sec:3.3} and  Cross Entropy simultaneously regulate local model to balance multiple domain performance.}}
}
\vspace{-10pt}
\label{fig:flowchart}
\end{figure}

\subsection{Discussion}
\label{sec:discussion}
\noindent{} \textbf{Conceptual Comparison}.
We illustrate the proposed method, \oursabbrv{} in  \cref{fig:framework} and \cref{alg:ours} and depict the flowchart in \cref{fig:flowchart}. Our method can be viewed as how to learn more information from others in collaborative updating and how to maintain multiple domains knowledge in local updating. Firstly, we perform communication at both logits and feature  levels via  \FCCMab{} (\cref{sec:3.1})  and \FISLab{} (\cref{sec:3.2}). Besides, it is applicable to large-scale scenario in federated learning, attributed to that the computation complexity is $\mathcal{O}(K)$. Notably, \FCCMab{} and \FISLab{} is respectively based on logits output and  similarity distribution regardless of the specific model structure. Thus, when participants share same model structure (model homogeneity), ours is still capable. We provide the comparison with state-of-the-art methods under model homogeneity in \cref{sec:comparHomo}. Secondly, in local updating, \FNTDab{} (\cref{sec:3.3}) disentangles logits knowledge distillation and focuses on transferring inter-domain knowledge, which does not incur heavy computational overhead and does not require complicated hyper-parameter.


\noindent{} \textbf{Generalization Bound}.
{\color{majorblack}{ Driven by the existing related analysis, including single-source domain and multi-domain learning \cite{AnaDA_NeurIPS07,LearnBoundforDA_NeurIPS07,DAwithMultipSourcesNeurIPS08,TheoryofLearnDiffDomains_ML10,AlgTheoryforMultipleSourceAda_NeurIPS18} we provide the insight into the generalization bound of heterogeneous federated learning. We denote the global distribution as $\mathcal{D}$, the $i$-th local distribution and its empirical distribution as $\mathcal{D}_i$ and $\hat{\mathcal{D}}_i$, respectively.
In our analysis, we assume a binary classification task, with hypothesis $h$ as a function $h: X \rightarrow \{0,1\}$. The task loss function is formulated as $l(h(x),y)=|\hat{y}-y|$, where $\hat{y}:=h(x)$. Note that
$l(\hat{y},y)$ is convex with respect to $\hat{y}$. We denote $\arg \min_{h \in \mathcal{H}} L_{\hat{\mathcal{D}}}(h)$ by $h_{\hat{\mathcal{D}}}$.}}
\begin{theorem} \label{thm:informal_risk_upper_bound_for_ensemble_model}
The hypothesis $h \in  \mathcal{H}$ learned on $\hat{\mathcal{D}}_i$ is denoted by $\smash{h_{\hat{\mathcal{D}}_i}}$.
The upper bound on the risk of $K$ local models on $\mathcal{D}$
mainly consists of two parts: 1)
the empirical risk of the model trained on the global empirical distribution $\hat{\mathcal{D}} = \frac{1}{K} \sum_i \hat{\mathcal{D}}_i$,
and 2) terms dependent on the distribution discrepancy between $\mathcal{D}_i$ and $\mathcal{D}$,
with the probability $1 - \delta$:
\begin{equation}\small
\setlength\abovedisplayskip{0pt} \setlength\belowdisplayskip{0pt}
	\begin{split}
        L_\mathcal{D} \Big( \frac{1}{K} \sum_i h_{\hat{\mathcal{D}}_i} \Big)
          & \leq
        L_{\hat{\mathcal{D}}} ( h_{\hat{\mathcal{D}}} )
        + \frac{1}{K} \sum_{i} \left( \frac{1}{2} d_{\mathcal{H} \Delta \mathcal{H}} (\mathcal{D}_i, \mathcal{D}) + \lambda_i \right)
         \\
        & + \frac{ 4 + \sqrt{ \log ( \tau_{\mathcal{H}} (2m) ) } }{ \delta / K \sqrt{ 2m } },
        \end{split}
\end{equation}
where $d_{\mathcal{H} \Delta \mathcal{H}}$ measures the distribution discrepancy between two distributions~\cite{ben2010theory},
$m$ is the number of samples per local distribution,
$\lambda_i$ is the minimum of the combined loss $\mathcal{L}_{\mathcal{D}} (h) \!+\! \mathcal{L}_{\mathcal{D}_i}(h), \forall h \in \mathcal{H}$,
and $\tau_{\mathcal{H}}$ is the growth function bounded by a polynomial of the VCdim of $\mathcal{H}$.
\end{theorem}
{\color{majorblack}
\noindent{} \textbf{Proof of \cref{thm:informal_risk_upper_bound_for_ensemble_model}}. We begin with the $L_\mathcal{D} \Big( \frac{1}{K} \sum_i h_{\hat{\mathcal{D}}_i} \Big)$ risk. By convexity of $l$ and Jensen inequality, we have: 
\begin{equation}\small
\setlength\abovedisplayskip{0pt} \setlength\belowdisplayskip{0pt}
        L_\mathcal{D} \Big( \frac{1}{K} \sum_i h_{\hat{\mathcal{D}}_i} \Big)
        \leq
     \frac{1}{K} \Big(\sum_i L_\mathcal{D}  h_{\hat{\mathcal{D}}_i} \Big).
\end{equation}
Via {\cite{TheoryofLearnDiffDomains_ML10,UnderstandML_14}}, we transfer from domain $D$ to $D_i$ as the following formulation:
\begin{equation}\small	
\begin{split}
\setlength\abovedisplayskip{0pt} \setlength\belowdisplayskip{0pt}
 L_\mathcal{D}  h_{\hat{\mathcal{D}}_i} &\leq  L_{\mathcal{D}_i} ( h_{\hat{\mathcal{D}}_i} ) + \frac{1}{2} d_{\mathcal{H} \Delta \mathcal{H}} (\mathcal{D}_i, \mathcal{D}) + \lambda_i,\\
 L_{\mathcal{D}_i} ( h_{\hat{\mathcal{D}}_i} ) &\leq  L_{\hat{\mathcal{D}}_i} ( h_{\hat{\mathcal{D}}_i} ) +  \frac{ 4 + \sqrt{ \log ( \tau_{\mathcal{H}} (2m) )}}{ \delta / K \sqrt{ 2m }}.
 \end{split}
\end{equation}
Based on the definition of ERM: $L_{\hat{\mathcal{D}}_i} ( h_{\hat{\mathcal{D}}_i} ) \leq L_{\hat{\mathcal{D}}_i} ( h_{\hat{\mathcal{D}}} )$ and the definition: $\hat{\mathcal{D}} = \frac{1}{K} \sum_i \hat{\mathcal{D}}_i$, we have:
\begin{equation}\small	
  \frac{1}{K} \sum_i L_{\hat{\mathcal{D}}_i} ( h_{\hat{\mathcal{D}}_i} ) \leq 
 \frac{1}{K} \sum_i  L_{\hat{\mathcal{D}}_i} ( h_{\hat{\mathcal{D}}} )=L_{\hat{\mathcal{D}}} ( h_{\hat{\mathcal{D}}} ) 
\end{equation}
\cref{thm:informal_risk_upper_bound_for_ensemble_model} reveals that
compared to the centralized model on the global empirical distribution, the performance on the global distribution is associated with the discrepancy between local distributions $\mathcal{D}_i$ and the global distribution $\mathcal{D}$. The results on different scenarios in \cref{table:interacc} confirm that on the relatively simple scenario, \ie, \digits{}, ours performs better improvement effect than other scenarios.
}

\noindent{} \textbf{Difference with Conference Version}.
Comparison with \fccl{} (\cite{FCCL_CVPR22}) (our conference paper),  it conducts  logits distillation in \cref{eq:kd_ori_form} with previous model ($\theta^{e-1}$) and pretrained model ($\theta^*$) on private data. The local loss is given as: 
\begin{equation}\small
\setlength\abovedisplayskip{0pt} \setlength\belowdisplayskip{0pt}
    \mathcal{L}_{{local}} = \mathcal{L}_{{CE}} + \mathcal{L}_{KD}(\mathcal{T}:\theta^{e-1})+\mathcal{L}_{KD}(\mathcal{T}:\theta^*).
    \label{eq:locafccl}
\end{equation}
Leveraging pre-trained model $\theta^*$ to do distillation ($\mathcal{L}_{KD}(\mathcal{T}:\theta^*)$) is to alleviate the gradient conflict mentioned in \cref{eq:grad_conflict}. However, a well-optimized pretrained model is the prerequisite, which can not be ensured for multiple participants. Therefore, this journal version introduces an improved version, namely \FNTDab{}, which simultaneously avoids incompatible optimization objective and gets rid of the assumption of requiring a powerful intra-domain teacher model. By \textbf{changing}
the $\mathcal{L}_{KD}(\mathcal{T}:\theta^{e-1})$ to $\mathcal{L}_{\FNTDab}(\mathcal{T}:\theta^{e-1})$, this already avoids the gradient conflict. Hence, the $\mathcal{L}_{KD}(\mathcal{T}:\theta^*)$ can be \textbf{deleted} in the local updating stage. Therefore, we design the novel and improved local updating objective in \cref{eq:localobj}, replacing the old one in \cref{eq:locafccl} as follows:
\begin{equation}\small
\setlength\abovedisplayskip{0pt} \setlength\belowdisplayskip{0pt}
    \mathcal{L}_{{local}} = \mathcal{L}_{{CE}} + \underbrace{\mathcal{L}_{KD}(\mathcal{T}:\theta^{e-1})}_{\Rightarrow \mathcal{L}_{\FNTDab}(\mathcal{T}:\theta^{e-1}) }+ \cancel{\mathcal{L}_{KD}(\mathcal{T}:\theta^*)}.
    \label{eq:localdiff}
\end{equation}

{
\color{majorblack}{
\noindent{} \textbf{Communication and Computation Discussion}.
Regarding the communication aspect, in the model heterogeneous federated learning, different clients hold distinct model architectures, rendering gradient/parameter averaging operations infeasible \cite{FLOptimization_arXiv16,FedAvg_AISTATS17,FedProx_MLSys2020,FedProc_arXiv21,MOON_CVPR21,FedLC_ICML22,FPL_CVPR23}. Consequently,when dealing with model heterogeneity, leveraging the output signals from the public data presents acts as a practicable communication strategy among heterogeneous clients. Closely related methods, \ie,  \fedmd{} \cite{FedMD_NeurIPS19} and \feddf{} \cite{FedDF_NeurIPS20} leverage the logits output to conduct the communication. We further conduct the \FISL{} module to achieve the feature-level communication for heterogeneous models in order to boost the generalizable performance. Thus, it avoidably brings additional communication cost: instance similarity matrix ($\mathcal{O}{(|B_0|^2)}$), which is dependent on the $|B_0|$ scale and does not linearly increase along the network parameter size. We summarize the collaborative communication cost for related methods in \cref{tab:communication}.
In terms of the local computation view, regularization signals act as a crucial character in calibrating the client optimizing direction as data heterogeneity introduces diverse distributions and thus distinct updating objectives. Thus, without regularization signals, clients would optimize towards the local minima and bring catastrophic forgetting on others knowledge, hindering the federated convergence speed. As for \oursabbrv{}, it utilizes the previous model to construct distillation to maintain the inter-domain performance, which is also a widely adopted signal to calibrate the local optimization {\cite{FedProx_MLSys2020,MOON_CVPR21,FedProc_arXiv21,FPL_CVPR23}}. We  provide the local computation cost comparison in the \cref{tab:computation}.
}}


\begin{table}[t]\small
\caption{
{\color{majorblack}\textbf{Communication cost} comparison with existing methods from  whether supports model heterogeneity (Model Heter.), the selection of communication signals (Comm. Signals), the degree of communication complexity (Comm. Cost). Experimental comparison on \digits{} inter-domain performance. - denotes these methods are not available for model heterogeneous settings. Refer to \cref{sec:discussion} and \cref{table:interacc}.}
}
\label{tab:communication}
\vspace{-10pt}
\centering
{
\resizebox{\columnwidth}{!}{
		\setlength\tabcolsep{1pt}
		\renewcommand\arraystretch{1.05}
\begin{tabular}{rIc|c|c|c}
\hline\thickhline
\rowcolor{mygray}
Methods & Comm. Signals & Comm. Cost  & Model Heter. & \digits{} Inter Acc.  \\
\hline\hline 
\fedavg{}\cite{FedAvg_AISTATS17} & Gradient & ${|\theta|}$  & \hlr{\xmark} & -  \\
\fedprox{}\cite{FedProx_MLSys2020} & Gradient   & ${|\theta|}$ & \hlr{\xmark} & - \\
\moon{}\cite{MOON_CVPR21}    & Gradient & ${|\theta|}$ & \hlr{\xmark} & - \\
\fedmd{}\cite{FedMD_NeurIPS19}   & Logits & ${|B_0| \!\times\! C}$  & \hlg{\tmark} & 26.38\\
\fccl{}\cite{FCCL_CVPR22}  & Logits & ${|B_0| \!\times\! C}$& \hlg{\tmark}  & 33.51 \\
\hline
\hline
\oursabbrv{} (Ours)  & Logits \&  Matrix  & ${|B_0| \!\times\! C \!+\! |B_0|^2}$  & \hlg{\tmark} & \textbf{50.49}
\end{tabular}}}
\vspace{-15pt}
\end{table}

\begin{table}[t]\small
\caption{
{\color{majorblack}\textbf{Local computation} comparison with existing methods from  the resource for parameter space (Para. Space.), the selection of regularization signals, and whether alleviate catastrophic forgetting (Cata. Forg.).  All methods default maintain an optimizing model. See details in \cref{sec:discussion}.}
}
\label{tab:computation}
\vspace{-10pt}
\centering
{
\resizebox{\columnwidth}{!}{
		\setlength\tabcolsep{1pt}
		\renewcommand\arraystretch{1.05}
\begin{tabular}{rIc|c|c}
\hline\thickhline
\rowcolor{mygray}
Methods & Para. Space   &  Regularization Signals & Cata. Forg. \\
\hline\hline 
\multicolumn{4}{l}{\textit{Only Model Homogeneous FL}} \\  \hline
\fedavg{}\cite{FedAvg_AISTATS17} & ${|\theta|}$ & - & \hlr{\xmark}    \\
\fedprox{}\cite{FedProx_MLSys2020} & ${2|\theta|}$  & Global Model & \hlg{\tmark} \\
\moon{}\cite{MOON_CVPR21} & ${3 |\theta|}$ & Global \& Previous Model  & \hlg{\tmark} \\
\hline\hline 
\multicolumn{4}{l}{\textit{Both Model Homogeneous and Heterogeneous FL}} \\  \hline
\fedmd{}\cite{FedMD_NeurIPS19} & ${|\theta|}$ & -  & \hlr{\xmark}  \\
\fccl{}\cite{FCCL_CVPR22}  & ${3 |\theta|}$ & Previous \& Pretrained Model   & \hlg{\tmark}\\
\hline
\hline
\oursabbrv{} (Ours)& ${2|\theta|}$   & Previous Model  & \hlg{\tmark}
\end{tabular}}}
\vspace{-15pt}
\end{table}

\begin{table}[t]\small
\caption{
{\color{majorblack}
\textbf{Computation analysis of proposed modules} from the increased communication cost (Comm. Cost) and Parameter Space (Para. Space.) aspects. See details in \cref{sec:discussion}.}
}
\label{tab:ourscomputation}
\vspace{-10pt}
\centering
{
\resizebox{\columnwidth}{!}{
		\setlength\tabcolsep{1pt}
		\renewcommand\arraystretch{1.05}
\begin{tabular}{rIc|c|c|c}
\hline\thickhline
\rowcolor{mygray}
Module  & Comm. Cost & Para. Space & Inter-Domain &  Intra-Domain \\
\hline\hline 
\multicolumn{3}{l|}{\textit{Collaborative Updating Stage}} & \multicolumn{2}{c}{{\digits{} Mean  Acc}}\\  \hline
\FCCMab{}\cref{sec:3.1}  & ${|B_0| \!\times\! C}$  & $|\theta|$ & 33.51 & 85.69\\
\FISLab{}{}\cref{sec:3.2}  & ${|B_0|^2}$ & $|\theta|$ & 37.74 & 90.14\\
\hline\hline 
\multicolumn{3}{l|}{\textit{Local Updating Stage}}& \multicolumn{2}{c}{{\digits{} Mean Acc}}\\  \hline
\FNTDab{}{}\cref{sec:3.3} & ${|B_i| \!\times\! C}$ & ${2|\theta|}$  & 50.49 & 90.57
\end{tabular}}}
\vspace{-15pt}
\end{table}

\section{Experiments}
\subsection{Experimental Setup}
\label{sec:ExperSetup}
\begin{figure}[h]
\centering
\vspace{-10pt}
\hspace{-6pt}
\subfigure[\digits{}]{
\includegraphics[width=0.24\linewidth]{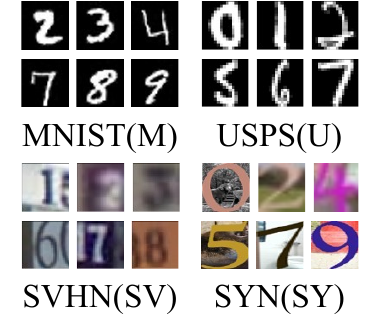}
}\hspace{-6pt}
\subfigure[\officecaltech{}]{
\includegraphics[width=0.24\linewidth]{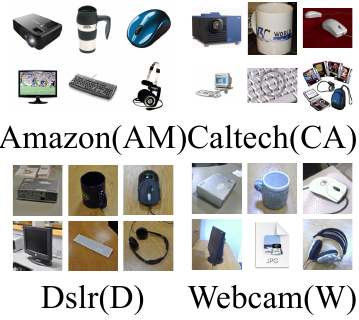}
}\hspace{-6pt}
\subfigure[\officeTO{}]{
\includegraphics[width=0.24\linewidth]{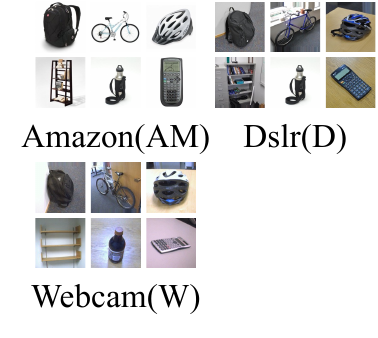}
}\hspace{-6pt}
\subfigure[\officehome{}]{
\includegraphics[width=0.24\linewidth]{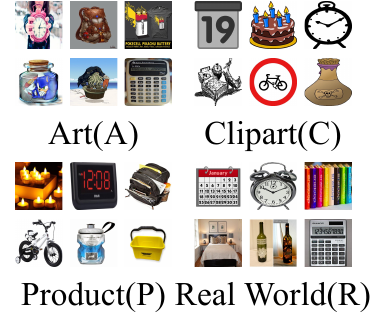}
}
\hspace{-6pt}
\vspace{-10pt}
\caption{\color{black}\textbf{Example cases} in \digits{}{\cite{MNIST_IEEE98,USPS_PAMI94,svhn_NeurIPS11,syn_arXiv18}}, \officecaltech{} {\cite{OffCaltech_CVPR12}}, \officeTO{} {\cite{office31_ECCV10}} and \officehome{} {\cite{officehome_CVPR17}} scenarios. See details in \cref{sec:ExperSetup}.}
\label{fig:example}
\vspace{-10pt}
\end{figure}

\noindent \textbf{Datasets and Models}.
We extensively evaluate our method on four classification scenarios (\ie, \digits{} {\cite{MNIST_IEEE98,USPS_PAMI94,svhn_NeurIPS11,syn_arXiv18}}, \officecaltech{} {\cite{OffCaltech_CVPR12}}, \officeTO {\cite{office31_ECCV10}} an \officehome {\cite{officehome_CVPR17}}) with three public data (\ie, \cifarhun{} {\cite{cifar_Toronto09}}, \tyimagenet{} {\cite{ImageNet_IJCV15}} and \fashionmnist {\cite{fashionmnist_arXiv17}}). 
For example, for \digits{} scenario, it includes four domains (\ie, \mnist{} ({M}), \usps{} ({U}), \svhn{} ({SV}) and \syn{} ({SY})) with $10$ categories. Note that for different scenarios, data acquired from different domains present domain shift (\textbf{data heterogeneity}). The example cases in each domain are presented in \cref{fig:example}. As shown in \cref{table:pridatascale}, we report the private data scale for each domain calculate and the proportion compared with the complete training data.
For these four classification scenarios, participants customize private models that have differentiated backbones and classifiers, (\textbf{model heterogeneity}). For experiments, we randomly set the backbone architecture as \resnet \cite{ResNet_CVPR16}, \efficientnet {\cite{EfficientNet_ICML19}}, \mobilenet {\cite{MobileNet_arXiv17}} and \googlenet {\cite{GoogLeNet_CVPR15}}. We summarize the selected ones for different domains in \cref{table:custmodel}. As for unified classifier, it maps the embedding feature layer to logits output with 10, 10, 31 and 65 dimensions, which is the classification categories of \digits{}, \officecaltech{}, \officeTO{} and \officehome{} scenarios, respectively.

\begin{table}[h]\small
\centering
\caption{\color{black}\textbf{Private data scale} in  different  federated scenarios. Refer to \cref{sec:ExperSetup}.}
\label{table:pridatascale}
\vspace{-10pt}
{
\resizebox{\linewidth}{!}{
\setlength\tabcolsep{1.5pt}
\begin{tabular}{c||ccccIcccc}
\hline\thickhline
\rowcolor{mygray}
& \multicolumn{4}{cI}{\digits}&\multicolumn{4}{c}{\officecaltech}\\
\cline{2-9}
\rowcolor{mygray}
\multirow{-2}{*}{Scale} & M & U & SV & SY 
& AM & CA  & D & W \\
\hline\hline
$N_i$&
\tworow{150}{0.25}&
\tworow{80}{1.10}&
\tworow{5000}{6.82}&
\tworow{1800}{18}&
\tworow{368}{48}&
\tworow{431}{48}&
\tworow{60}{48}&
\tworow{113}{48}\\

\hline\thickhline
\rowcolor{mygray}
& \multicolumn{4}{cI}{\officeTO}&\multicolumn{4}{c}{\officehome}\\
\cline{2-9}
\rowcolor{mygray}
\multirow{-2}{*}{Scale} & AM & D & W & 
& A & C & P & R\\

\hline\hline
$N_i$&
\tworow{1082}{48}&
\tworow{192}{48}&
\tworow{305}{48}&&
\tworow{1400}{82.35}&
\tworow{2000}{65.57}&
\tworow{2500}{80.65}&
\tworow{2000}{65.57}
\end{tabular}}}	
\vspace{-10pt}
\end{table}

\begin{table}[t]\small
\centering
\caption{\color{black}\textbf{Customized model} in  four federated scenarios. See details in \cref{sec:ExperSetup}.}
\label{table:custmodel}
\vspace{-10pt}
{
\resizebox{\linewidth}{!}{
\setlength\tabcolsep{1.5pt}
\begin{tabular}{ccccIcccc}
\hline\thickhline
\rowcolor{mygray}
\multicolumn{4}{cI}{\digits}&\multicolumn{4}{c}{\officecaltech}\\
\cline{1-8}
\rowcolor{mygray}
M & U & SV & SY 
& AM & CA  & D & W \\
\hline\hline

\resnetten & \resnettwelve & \efficientnet & \mobilenet 
& \googlenet & \resnettwelve & \resnetten & \resnettwelve \\

\hline\thickhline
\rowcolor{mygray}
\multicolumn{4}{cI}{\officeTO}&\multicolumn{4}{c}{\officehome}\\
\cline{1-8}
\rowcolor{mygray}
AM & D & W & 
& A & C & P & R\\
\hline\hline
\resnetten & \resnettwelve & \resnetten &  
& \resneteighteen & \resnetthirtyfour & \googlenet & \resnettwelve 
\end{tabular}}}	
\vspace{-10pt}
\end{table}

\begin{table}[t]\small
\centering
\caption{\color{black}{\textbf{Hyper-parameters for different federated methods} in \digits{} scenario. Please see more details in \cref{sec:ExperSetup}.}}
\label{tab:chosen_hyper_param_table}
\vspace{-10pt}
\resizebox{\columnwidth}{!}{
\begin{tabular}{r|l}
\hline\thickhline
\rowcolor{mygray}
{Method}  & {Parameters} \\
\hline\hline
\fedprox{} \cite{FedProx_MLSys2020}
& $\mu$: 0.01\\
\moon{} \cite{MOON_CVPR21}
& $\tau$: 0.5, $\mu$: 1\\
\fedproc{} \cite{FedProc_arXiv21}
& $\alpha$: 0.5  \\
\fedrs{}{} \cite{FedRS_KDD21}
& $\alpha$: 0.5\\
\rhfl{} \cite{RHFL_CVPR22}
& $\alpha$: 0.1,  $\beta$: 1.0\\
\fccl{} \cite{FCCL_CVPR22}
& $\lambda$: 0.0051 \\
\hline
\oursabbrv{} 
& $\mu$: 0.002, $\omega$: 3 in \cref{sec:ablation_frsl}, $\tau$: 3 in \cref{sec:ablation_fntd} \\
\end{tabular}
}
\vspace{-10pt}
\end{table} 

\noindent \textbf{Evaluation Metric}.
We report the standard metrics to measure the quality of methods: accuracy, which is defined as the number of samples that are paired divided by the number of samples. Specifically, for evaluation intra and inter-domain performance, we define as follows:
\begin{equation}\small 
\setlength\abovedisplayskip{0pt} \setlength\belowdisplayskip{0pt}
\mathcal{A}_i^{Intra}=\frac{\sum(\texttt{argMax}(f(\theta_i,X_i^{Test}))==Y_i^{Test})}{|D_i^{Test}|},
\label{eq:intraAcc}
\end{equation}
\begin{equation}\small 
\setlength\abovedisplayskip{0pt} \setlength\belowdisplayskip{0pt}
\mathcal{A}_i^{Inter}=\sum_{j \neq i}\frac{\sum(\texttt{argMax}(f(\theta_i,X_j^{Test}))==Y_j^{Test})}{(K-1)\times|D_j^{Test}|}.
\label{eq:interAcc}
\end{equation}
For overall performance evaluation, we adopt the average accuracy. Besides, \digits{}, \officecaltech{}, \officeTO{} and \officehome{} contain 10, 10, 31 and 65 categories. Top-1 and Top-5 accuracy are adopted for the first two and the latter two.

\noindent \textbf{Counterparts}.
{\color{black}{
We compare relative federated methods: 

\noindent {- For model heterogeneity based:}
\begin{itemize} 
\item \fedmd{} \pub{NeurIPS'19} {\cite{FedMD_NeurIPS19}}: Rely on the related public dataset to conduct distillation operation.
\item \feddf{} \pub{NeurIPS'19} {\cite{FedDF_NeurIPS20}}: Ensemble knowledge distillation for robust model fusion via unlabeled data.
\item \fedmatch{} \pub{ICLR'21} {\cite{FedMatch_ICLR21}}: Inter-client consistency and disjoint parameters for disjoint learning. 
\item \rhfl{} \pub{CVPR'22} \cite{RHFL_CVPR22}: Robust noise-tolerant loss for heterogeneous clients under label noise federated learning
\item \fccl{} \pub{CVPR'22} {\cite{FCCL_CVPR22}}: Align Cross-Correlation with unlabeled data and distill inter-domain knowledge
\end{itemize}

\noindent {- For model homogeneity based:}
\begin{itemize} 
\item \fedavg{}{} \pub{AISTATS'17} {\cite{FedAvg_AISTATS17}}: Average model parameter to realize communication in federated learning

\item \fedprox{}{} \pub{MLSys'20} {\cite{FedProx_MLSys2020}}: Introduce proximal term with the global model parameter via the $\mathcal{L}_2$ function

\item \moon{}{} \pub{CVPR'21} {\cite{MOON_CVPR21}}: Pull to the global model and push away the previous local model on feature level
\item \fedproc{}{} \pub{arXiv'21} {\cite{FedProc_arXiv21}}: Leverage global prototypes to conduct prototype-level contrastive learning
\item  \fedrs{} \pub{SIGKDD'21} {\cite{FedRS_KDD21}}: “Restricted Softmax" limits the update of missing classes’ weights
\end{itemize}
}}


\begin{table*}[t]\small
\centering
\caption{\color{black}\textbf{Comparison of inter-domain generalization ($\rightarrow$ ) performance} with counterparts on four scenarios with \cifarhun{}. {M}$\rightarrow$ means that private data is \mnist{} and the respective model is tested on all other domains except itself in \cref{eq:interAcc}. AVG denotes the average inter-domain accuracy calculated from each domain. We measure the performance via the mean accuracy of the last three communication epochs. See details in \cref{sec:comparSOTA}. Best in  \textbf{bold} and second with \underline{underline}. These notes are the same to others.}
\label{table:interacc}
\vspace{-10pt}
{
\resizebox{0.9\textwidth}{!}{
\begin{tabular}{r||ccccIccIccccIcc}
\hline\thickhline
\rowcolor{mygray}
& \multicolumn{6}{cI}{\digits}&\multicolumn{6}{c}{\officecaltech}\\
\cline{2-13}
\rowcolor{mygray}
\multirow{-2}{*}{Methods} & M$\rightarrow$ & U$\rightarrow$ & SV$\rightarrow$ & SY$\rightarrow$ & AVG & $\triangle$
& AM$\rightarrow$ & CA$\rightarrow$ & D$\rightarrow$ & W$\rightarrow$ & AVG & $\triangle$ \\

\hline\hline

\solo 
& 15.29 & 13.91 & 39.24 & 34.30 & 25.68 & -
& 28.20 & 32.11 & 19.99 & 28.11 & 27.10 & -\\     

\fedmd {\cite{FedMD_NeurIPS19}} 
&  8.97 & 12.61 & 40.89 & 43.03 & 26.38 & +0.70
& 21.09 & 35.13 & 21.76 & 30.57 & 27.13 & +0.03\\

\feddf {\cite{FedDF_NeurIPS20}} 
& 13.23 & 19.29 & 45.25 & 43.95 & 30.43 & +4.75
& 23.87 & 28.29 & 16.27 & 22.82 & 22.81 & {-4.29}\\
          
\fedmatch {\cite{FedMatch_ICLR21}} 
&  9.22 & 14.76 & 46.28 & 36.05 & 26.58 & +0.90
& 19.25 & 32.21 & 13.80 & 22.66 & 21.98 & -5.12\\

\rhfl{} \cite{RHFL_CVPR22} 
& 19.15 & 16.72 & \textbf{51.74} & \underline{48.65} & \underline{34.06} & +8.38
& 19.11 & 27.50 & 16.55 & 23.83 & 21.74 & -5.36\\

\hline

Our \fccl{} \cite{FCCL_CVPR22} 
&\underline{20.74} & \underline{20.60} & 44.68& {48.02} & {33.51} & \underline{+7.83} 
& \underline{25.16} & \underline{33.68} & \underline{17.52} & 23.81 & \underline{25.04} & -3.07 \\

Our \oursabbrv{} 
& \textbf{40.13} & \textbf{50.53} & \underline{48.31} & \textbf{63.00} & \textbf{50.49} & \textbf{+24.81}
& \textbf{29.56} & \textbf{36.82} & \textbf{24.02} & \textbf{31.33} & \textbf{30.43} & \textbf{+3.33}\\

\hline\thickhline
\rowcolor{mygray}
& \multicolumn{6}{cI}{\officeTO}&\multicolumn{6}{c}{\officehome}\\
\cline{2-13}
\rowcolor{mygray}
\multirow{-2}{*}{Methods} & {AM}$\rightarrow$ & {D}$\rightarrow$ & {W}$\rightarrow$ & & AVG & $\triangle$ &{A}$\rightarrow$  & {C}$\rightarrow$  & {P}$\rightarrow$  & {R}$\rightarrow$  & AVG & $\triangle$\\

\hline\hline

\solo 
& 21.08 & 27.93 & 34.14 & & 27.72 & -
& 18.89 & 19.36 & 21.97 & 21.02 & 20.31 & - \\

\fedmd {\cite{FedMD_NeurIPS19}} 
& 26.26 & 32.37 & 40.58 &  &33.07 & +5.35
& 16.85 & 23.13 & 28.78 & 25.01 & 23.44 & +3.13\\

\feddf {\cite{FedDF_NeurIPS20}} 
& 24.74 & 26.70 & 35.40 &  & 28.96 & +1.24
& 17.38 & 21.76 & 25.17 & 22.97 & 21.82 & +1.51\\

\fedmatch {\cite{FedMatch_ICLR21}} 
& 22.23 & 31.98 & 40.15 &       & 31.45 & +3.73
& 19.05 & 25.24 & \underline{28.73} & 24.35 & 24.34 & +4.03\\

\rhfl{} \cite{RHFL_CVPR22} 
& 20.52 & 20.59 & 36.32 & & 25.81 & -1.91
& 15.67 & 20.70 & 22.02 & 26.06 & 21.11 & +0.80\\

\hline

Our \fccl{} \cite{FCCL_CVPR22} 
& \underline{26.69} &  \underline{34.01} & \underline{39.88} & & \underline{33.52} & \underline{+5.80}
& \underline{25.55} &  \textbf{26.41}  &  \textbf{30.14}  &   \underline{29.41}  & \underline{27.88} & \underline{+7.49}\\

Our \oursabbrv{} 
& \textbf{31.87} & \textbf{35.16} & \textbf{44.62} && \textbf{37.21} & \textbf{+9.49}
& \textbf{26.67} & \underline{26.07} & 25.96 & \textbf{33.02} & \textbf{27.93} &\textbf{+7.62}\\
\end{tabular}}}
\vspace{-10pt}
\end{table*}

\begin{table*}[t]\small
\centering
\caption{\textbf{Comparison of intra-domain ($\hookleftarrow$) performance} with counterparts on four scenarios with \cifarhun. The intra-domain accuracy metric ($\hookleftarrow$) in \cref{eq:intraAcc} is evaluated on respective testing data. See details in \cref{sec:comparSOTA}.}
\label{table:intraacc}
\vspace{-10pt}
{
\resizebox{0.9\textwidth}{!}{
\begin{tabular}{r||ccccIccIccccIcc}
\hline \thickhline
\rowcolor{mygray}
& \multicolumn{6}{cI}{\digits}&\multicolumn{6}{c}{\officecaltech}\\
\cline{2-13} 
\rowcolor{mygray}
\multirow{-2}{*}{Methods}  
& M$\hookleftarrow$ & U$\hookleftarrow$ & SV$\hookleftarrow$ & SY$\hookleftarrow$ & AVG & $\triangle$ 
& AM$\hookleftarrow$ & CA$\hookleftarrow$ & D$\hookleftarrow$ & W$\hookleftarrow$ & AVG & $\triangle$ \\

\hline\hline

\solo 
& 70.20 & 74.19 & 74.57 & 73.60 & 73.14 & -
& 74.11 & \textbf{61.35} & \underline{75.16} & \textbf{78.31} & \textbf{72.23} & -\\

\fedmd {\cite{FedMD_NeurIPS19}} 
& 77.30 & 80.05 & 77.73 & 87.72 & 80.70 & +7.56
& 71.78 & 57.29 & 68.37 & 72.20 & 67.41 & -4.82\\

\feddf {\cite{FedDF_NeurIPS20}}  
& 82.95 & 78.84 & 78.46 & 91.30 & 83.38 & +10.24
& 70.01 & 56.49 & 65.18 & 64.86 & 64.13 & -8.10\\

\fedmatch {\cite{FedMatch_ICLR21}} 
& 82.69 & 78.31 & 79.79 & 89.23 & 82.69 & +9.55
& 73.38 & {60.25} & 68.58 & 73.56 & 68.94 & -3.29\\

\rhfl{} \cite{RHFL_CVPR22} 
& 86.69 & 80.39 & \textbf{88.44} & \textbf{97.90} & \underline{88.35} & \underline{+15.21}
& 73.35 & 58.92 & 70.91 & {74.47} & 69.41 & -2.82 \\

\hline

Our  \fccl{} \cite{FCCL_CVPR22} 
& \underline{88.84} & \underline{84.42} & 78.55 & 91.23 & 85.69 & +12.55
& \underline{75.09} & \underline{60.46} & {74.95} & 73.56 & {71.01} & {-1.22}\\

Our \oursabbrv{} 
& \textbf{90.54} & \textbf{86.92} & \underline{87.11} & \underline{97.72} & \textbf{90.57} & \textbf{+17.43}
& \textbf{75.26} & {59.90} & \textbf{75.80} & \underline{75.48} & \underline{71.60} & {-0.63}\\

\hline \thickhline
\rowcolor{mygray}
& \multicolumn{6}{cI}{\officeTO}&\multicolumn{6}{c}{\officehome}\\
\cline{2-13}
\rowcolor{mygray}
\multirow{-2}{*}{Methods} & AM$\hookleftarrow$ & D$\hookleftarrow$ & W$\hookleftarrow$ & & AVG & $\triangle$ 
& A$\hookleftarrow$ & C$\hookleftarrow$ & P$\hookleftarrow$ & R$\hookleftarrow$ & AVG & $\triangle$ \\

\hline \hline

\solo 
& \textbf{72.95} & 77.31 & 80.88 & & 77.04 & -
& 65.27 & 60.50 & 74.68 & 54.28 & 63.68 & -\\

\fedmd {\cite{FedMD_NeurIPS19}} 
& 71.39 & 76.24 & 80.13 & & 75.92 & -1.12
& 66.17 & 60.63 & 76.35 & 56.60 & 64.93 & +1.25\\

\feddf {\cite{FedDF_NeurIPS20}} 
& 72.25 & 74.50 & 80.34 & & 75.69 & -1.35
& 66.10 & 60.44 & 75.70 & 55.98 & 64.55 & +0.87\\

\fedmatch {\cite{FedMatch_ICLR21}} 
& 71.08 & 76.04 & 80.92 & & 76.01 & -1.03   
& \underline{81.50} & 65.40 & \underline{79.81} & \underline{65.06} & \underline{72.94} & \underline{+9.26}\\

\rhfl{} \cite{RHFL_CVPR22} 
& 70.98 & 73.90 & 78.87  & & 74.58 & -2.46 
& 56.68 & \underline{73.79} & 76.29 & 63.63 & 67.59 & +3.91 \\

\hline

Our \fccl{} \cite{FCCL_CVPR22} 
& {72.37} & \underline{78.44} & \textbf{81.26} & & \underline{77.35} & \underline{+0.31}
& \textbf{81.51} &  65.42 &  \textbf{79.84}  & \underline{65.16} & \textbf{72.98} & \textbf{+9.30}\\

Our \oursabbrv{} 
& \underline{72.92} & \textbf{78.61} & \textbf{81.26} & & \textbf{77.59} & \textbf{+0.55}
& 69.74 & \textbf{74.65} & 78.02 & \textbf{69.02} & 72.85 & {+9.17}\\
\end{tabular}}}
\vspace{-15pt}
\end{table*}

\begin{figure*}[t]
\centering
\subfigure[\digits{}]{
\includegraphics[width=0.47\textwidth]{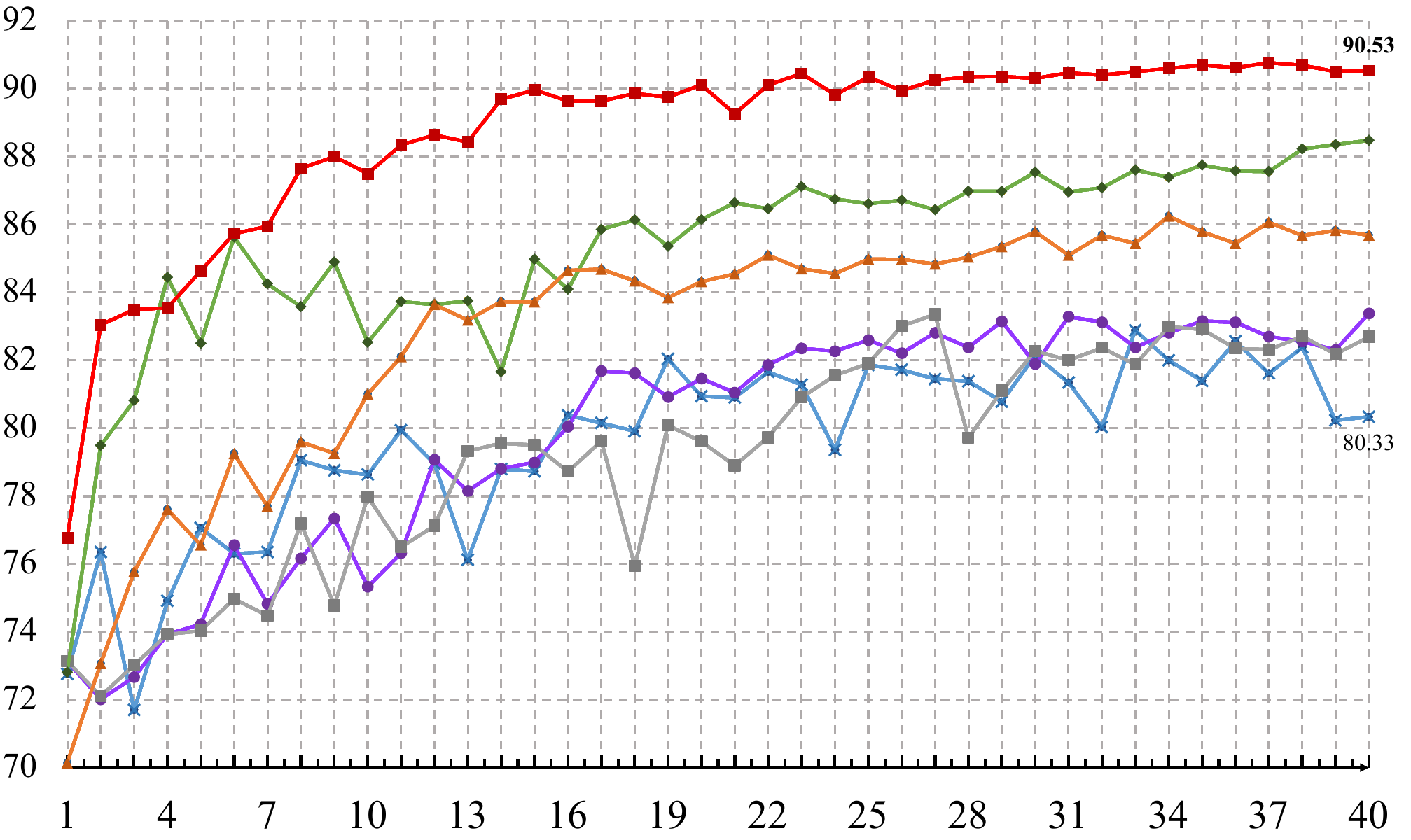}}
\subfigure[\officecaltech{}]{
\includegraphics[width=0.47\textwidth]{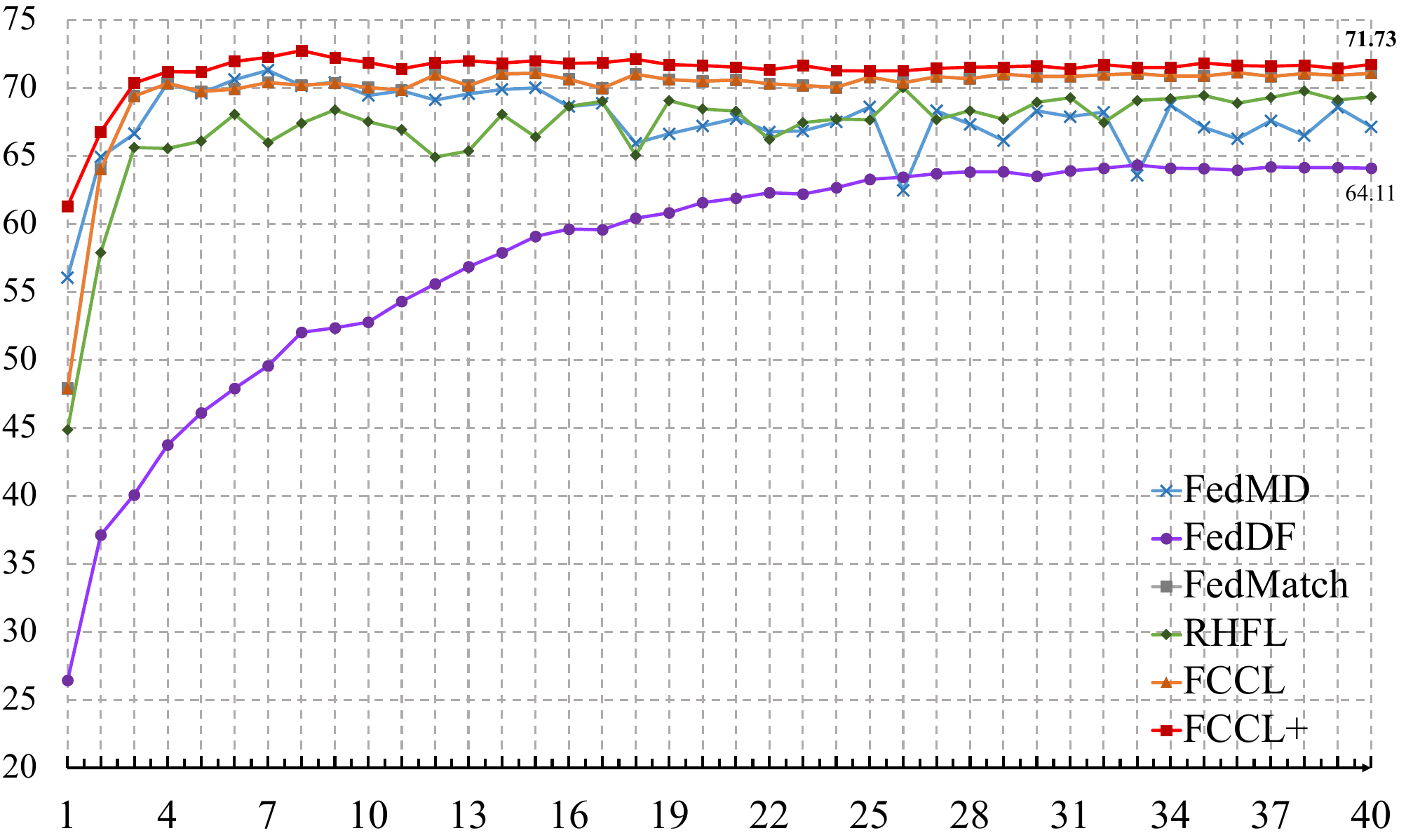}}
\vspace{-10pt}
\caption{\color{black}\textbf{Comparison of average intra-domain accuracy} during federated learning process with counterparts  in \digits{} and \officecaltech{} scenarios with \cifarhun{}. The horizontal and vertical coordinates represent the communication epoch and intra-domain accuracy. We mark the best and worst performance of the last communication epoch. Refer to \cref{sec:comparSOTA} for details.}
\label{fig:intrawithSOTA}
\vspace{-10pt}
\end{figure*}

\noindent \textbf{Implement Details}.
For a fair comparison, we follow {\cite{FedDF_NeurIPS20,MOON_CVPR21,FCCL_CVPR22,RHFL_CVPR22}}. The clients scale are dependent on the experimental scenario. For example, in digits, there are four clients. Models are trained via Adam optimizer {\cite{Adam_arXiv14}} with batch size of $512$ and $256$ in collaborative updating and local updating. The learning rate is consistent $0.001$ in these two process for all approaches. We set the hyper-parameter  $\lambda=0.0051$ like {\cite{Barlow_ICML21}} and $\mu=0.02$ in {\cite{ISD_ICCV21}}). We further conduct hyper-parameter analysis of $\omega$ and $\tau$ in \cref{table:omegaabaltion} and \cref{fig:ab_fntd}.
The number of unlabeled public data is $5000$ for different scenarios. Besides, we analyze the effect of strong and weak data augmentations on public data for \FISLab{} in \cref{sec:ablation_frsl}, and follow the data augmentation in \cite{SimCLR_ICML20,SCL_NeurIPS20} to construct \textbf{strong} data augmentation We list these data augmentations  following the PyTorch notations:

\begin{itemize}[leftmargin=*]
	\setlength{\itemsep}{0pt}
	\setlength{\parsep}{-2pt}
	\setlength{\parskip}{-0pt}
	\setlength{\leftmargin}{-10pt}
	\item \texttt{RandomResizedCrop}: Crop  with the scale of $(0.2, 1.)$. and images are resized to $32 \times 32$.
	\item \texttt{RandomHorizontalFlip}: The given image is horizontally flipped randomly with a given probability $p=0.5$.
	\item \texttt{ColorJitter}: We randomly change  the brightness, contrast, saturation and hue of images with value as $(0.4, 0.4,0.4,$ $ 0.1)$ and set the probability as $0.8$.
	\item \texttt{RandomGrayscale}: Randomly convert image to grayscale with a probability of $p=0.2$.
\end{itemize}

{\color{black}{
Besides, for the \textbf{weak} data augmentation, we only maintain \texttt{RandomResizedCrop} and \texttt{RandomHorizontalFlip} strategies for public data. See detail in \cref{sec:ExperSetup} and relative analysis in \cref{sec:ablation_frsl}.
We follow the data augmentation strategy in \cite{SimCLR_ICML20,SCL_NeurIPS20} to construct \textbf{strong} data augmentation. We list these data augmentations  following the PyTorch notations:
\begin{itemize} 
\item \texttt{RandomResizedCrop}: We crop \cifarhun{}, \tyimagenet{} and with the scale of $(0.2, 1.)$. The cropped images are resized to $32 \times 32$, $32 \times 32$ for these two public data.
\item \texttt{RandomHorizontalFlip}: The given image is horizontally flipped randomly with a given probability $p=0.5$.
\item \texttt{ColorJitter}: We randomly change  the brightness, contrast, saturation and hue of images with value as $(0.4, 0.4,0.4,$ $ 0.1)$ and set the probability as $0.8$.
\item \texttt{RandomGrayscale}: Randomly convert image to grayscale with a probability of $p=0.2$.
\end{itemize}
For pre-processing, we resize all input images into $32 \times 32$ with three channels for compatibility.  We do the communication for $E\!=\!40$ epochs, where all approaches have little or no accuracy gain with more rounds. For \solo{}, models are optimized on private data for $50$ epochs.}}
\subsection{Comparison with SOTA Methods}
\label{sec:comparSOTA}
We comprehensively examine the proposed method with state-of-the-art methods on four image classification tasks, \ie, \digits{} {\cite{MNIST_IEEE98,USPS_PAMI94,svhn_NeurIPS11,syn_arXiv18}}, \officecaltech{} {\cite{OffCaltech_CVPR12}}, \officeTO{} {\cite{office31_ECCV10}} and \officehome{} {\cite{officehome_CVPR17}}) with three public data (\ie, \cifarhun{} \cite{cifar_Toronto09}, \tyimagenet{} \cite{ImageNet_IJCV15} and \fashionmnist{} {\cite{fashionmnist_arXiv17}}).

\noindent \textbf{Inter-domain Generalization Analysis}.  
We summarize the results of inter-domain accuracy by the end of federated learning in \cref{table:interacc}. As shown in the table, it clearly depicts that under domain shift, \solo{} presents worst in these two tasks, demonstrating the benefits of federated learning. \oursabbrv{} consistently outperforms all other counterparts on different scenarios. For example, in \digits{} scenario, \oursabbrv{}  outperforms the best counterparts by  $19.39\%$ on \mnist{}, which is calculated through the average testing accuracy for respective model on \usps{}, \syn{} and \svhn{} testing data. 
{\color{black}
We conduct experiments in \cref{table:digitswithTYandFMNIST} with the \tyimagenet{} datasets on the \digits{} scenario, and ours  achieves consistently superior performance.}

\noindent \textbf{Intra-domain Discrimination Analysis}.
{\color{majorblack}
To demonstrate the effectiveness of alleviating catastrophic forgetting on intra-domain, we observe that our method almost outperforms the compared methods by a large margin along the federated learning process in \cref{fig:intrawithSOTA}. We further report the final intra-domain accuracy in \cref{table:intraacc}.  Although \solo{} achieves relatively competitive intra-domain performance, the reason is that it purely optimizes on the local data without collaborating with others.  However, the lack of collaboration hinders the generalization  of \solo{}  on inter-domain evaluation, where the data distributions may be significantly distinct from the local data.
\oursabbrv{} suffers less periodic performance shock and is not prone to overfitting to local data distribution. The  \cref{table:interacc}  and \cref{table:intraacc} illustrate that \oursabbrv{} is capable of balancing multiple domain knowledge and effectively alleviates the catastrophic forgetting problem. 
}

\begin{table*}[t]\small
\centering
\caption{\textbf{Ablation study of key components} of our proposed method in \digits{} and \officeTO{} scenarios with \cifarhun{}. Both inter-domain generalization ($\rightarrow$) and intra-domain discriminability ($\hookleftarrow$) accuracies are reported. Please see \cref{sec:ablation} for details.}
\label{table:ablation}
\vspace{-10pt}
{
\resizebox{0.9\textwidth}{!}{
\begin{tabular}{ccc||ccccIccIcccIcc}
\hline \thickhline
\rowcolor{mygray}
&&& \multicolumn{6}{cI}{\digits}&\multicolumn{5}{c}{\officeTO{}}\\
\cline{4-14} 
\rowcolor{mygray}
\multirow{-2}{*}{\FCCMab{}}& \multirow{-2}{*}{\FISLab{}} & \multirow{-2}{*}{\FNTDab{}} 
& M$\rightarrow$ & U$\rightarrow$ & SV$\rightarrow$ & SY$\rightarrow$ & AVG & $\triangle$
& AM$\rightarrow$ & D$\rightarrow$ & W$\rightarrow$ & AVG & $\triangle$\\

\hline\hline

& & 
& 15.29 & 13.91 & 39.24 & 34.30 & 26.68 & -
& 21.08 & 27.93 & 34.14 & 27.72 & -\\

\ding{51} & & 
& 20.74 & 20.60 & 44.68 & 48.02 & 33.51 & +6.83
&{26.69} & {34.01} & {39.88} & {33.52} & +5.80\\

& \ding{51} & 
& 25.59 & 24.75 & 47.53 & 53.12 & 37.74 & +11.06
& 20.02 & 28.04 & 36.51 & 28.19 & +0.47\\

\ding{51} & \ding{51}&
& 30.97 & 25.90 & \textbf{51.12} & {53.84} & 40.45 & {+13.77}
& 26.08 & 31.64 & 38.74 & 32.15 & +4.43\\

\ding{51} & \ding{51}& \ding{51}
& \textbf{40.13} & \textbf{50.53} & 48.31 & \textbf{63.00} & \textbf{50.49}& \textbf{+23.81}
& \textbf{30.95} & \textbf{32.77 } & \textbf{45.38} & \textbf{36.36} & \textbf{+8.64} \\
\hline \thickhline
\rowcolor{mygray}
&&& \multicolumn{6}{cI}{\digits}&\multicolumn{5}{c}{\officeTO}\\
\cline{4-14} 
\rowcolor{mygray}
\multirow{-2}{*}{\FCCMab{}}& \multirow{-2}{*}{\FISLab{}} & \multirow{-2}{*}{\FNTDab{}} 
& M$\hookleftarrow$ & U$\hookleftarrow$ & SV$\hookleftarrow$ & SY$\hookleftarrow$ & AVG & $\triangle$ 
& AM$\hookleftarrow$ & D$\hookleftarrow$ & W$\hookleftarrow$ & AVG & $\triangle$\\

\hline\hline

& & 
& 70.20 & 74.19 & 74.57 & 73.60 & 73.14  & - 
& \textbf{72.95} & 77.31 & 80.88 & 77.04 & - \\

\ding{51}& & 
& 88.84 & 84.42 & 78.55 & 91.23 & 85.69 & +12.55
&{72.37} & {78.44} & {81.26} & {77.35} & +0.31 \\

& \ding{51} & 
& 88.36 & 86.68 & \textbf{88.42} & 97.12 & 90.14 & +17.00
& 72.32 & 77.91 & 81.59 & 77.27 & +0.23\\

\ding{51} & \ding{51} & 
& 88.27 & 87.81 & 88.04 & 97.08 & 90.30 & +17.16
& 72.79 & 77.91 & \textbf{81.81} & 77.50  & +0.46\\

\ding{51} & \ding{51} & \ding{51}
& \textbf{90.54} & \textbf{86.92} & 87.11 & \textbf{97.72} & \textbf{90.57} & \textbf{+17.43}
& {72.92} & \textbf{78.61} & {81.26} & \textbf{77.59} & \textbf{+0.51}\\
\end{tabular}}}
\vspace{-15pt}
\end{table*}

\begin{table}[t]
\centering
\caption{\color{black}{\textbf{Analysis of \FCCMab{} with different hyper-parameter $\lambda$ in \cref{eq:fccm_loss}} on the \officeTO{} scenario for the inter- and intra- overall performance. See details in \cref{sec:ablation_fntd}.}}
\label{tab:analysisoflamdainFCCM}
\vspace{-10pt}
\centering
{
\resizebox{\linewidth}{!}{
\setlength\tabcolsep{1.5pt}
\begin{tabular}{cccccIccccc}
\hline \thickhline
\rowcolor{mygray}
\multicolumn{5}{cI}{Inter-domain} 
& \multicolumn{5}{c}{Intra-domain} \\
\hline
\rowcolor{mygray} 
0.002 & 0.0051 & 0.01 & 0.05 & 0.1 & 
0.002 & 0.0051 & 0.01 & 0.05 & 0.1
\\
\hline \hline
33.09 & 33.52 & 32.89 & 32.41 & 32.51 & 
77.54 & 77.35 & 77.60 & 77.00 & 77.53
\end{tabular}}}
\vspace{-10pt}
\end{table}

\begin{figure}[t]
\begin{minipage}[]{0.56\linewidth}
\centering
	\includegraphics[width=\linewidth]{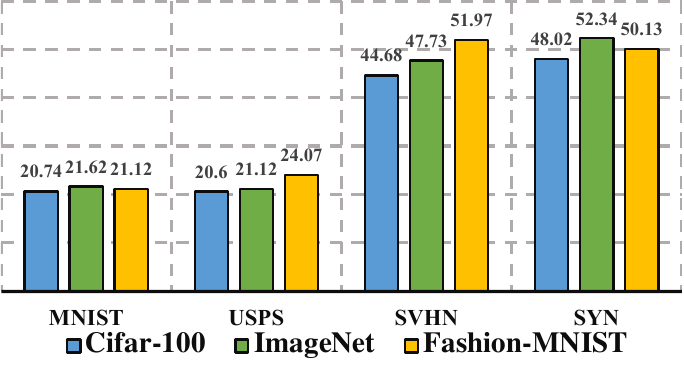}
\end{minipage}
\hfill
\begin{minipage}[]{0.42\linewidth}
\centering
\small{
\resizebox{\linewidth}{!}{
	\renewcommand\arraystretch{1.3}
	\begin{tabular}{r||c}
	\hline\thickhline
	\rowcolor{mygray}
	Public Data& AVG\\
    \hline\hline
    \cifarhun&33.51 \\
    \tyimagenet&35.70\\
    \fashionmnist& 36.82\\
	\end{tabular}
}}
\end{minipage}
\vspace{-5pt}
\caption{\textbf{Ablation study on \FCCMab{}} (\cref{sec:3.1}) with \textbf{different public data} for inter-domain generalization ($\rightarrow$) on each domain (\textbf{Histogram}) and overall performance (\textbf{Table}) in \digits{} task. See details in \cref{sec:ablation_fccm}.}
\label{fig:ab_fcc_inter}
\vspace{-10pt}
\end{figure}

\begin{table}[h]\small
\centering
\caption{\textbf{Comparison of inter-domain generalization ($\rightarrow$ ) performance and intra-domain discriminability ($\hookleftarrow$)} with counterparts on \digits{} scenario with \tyimagenet{}. "/" means optimization dilemma, \ie, \feddf{} with \tyimagenet{}. Refer to \cref{sec:comparSOTA}.}
\label{table:digitswithTYandFMNIST}
\vspace{-10pt}
{
\resizebox{0.95\columnwidth}{!}{
\begin{tabular}{r||ccccIc}
\hline\thickhline
\rowcolor{mygray}
& \multicolumn{5}{c}{\tyimagenet{}}\\
\cline{2-5}
\rowcolor{mygray}
\multirow{-2}{*}{Methods} 
& M$\rightarrow$ & U$\rightarrow$ & SV$\rightarrow$ & SY$\rightarrow$ & AVG \\

\hline\hline

\solo 
& 15.29 & 13.91 & 39.24 & 34.30 & 25.68 
\\

\fedmd {\cite{FedMD_NeurIPS19}} 
& 18.50 & 13.26 & 51.23 & 52.06 & 33.76
\\

\fedmatch {\cite{FedMatch_ICLR21}} 
& 11.77 & 11.49 & 46.65 & 46.33 & 29.06 
\\

\rhfl{} \cite{RHFL_CVPR22} 
& 16.14 & 14.05 & \textbf{54.26} & 47.57 & 33.00
\\

\hline

Our \fccl{} \cite{FCCL_CVPR22} 
& \textbf{31.84} & \underline{31.03} & 46.81 & \underline{52.75} & \underline{40.60} 
\\

Our \oursabbrv{} 
& \underline{30.15} & \textbf{48.33} & \underline{50.88} & \textbf{61.84} & \textbf{47.80} 
\\

\hline\thickhline
\rowcolor{mygray}
& \multicolumn{5}{c}{\tyimagenet{}}\\
\cline{2-5}
\rowcolor{mygray}
\multirow{-2}{*}{Methods} 
& M$\hookleftarrow$ & U$\hookleftarrow$ & SV$\hookleftarrow$ & SY$\hookleftarrow$ & AVG 
\\

\hline\hline

\solo 
& 70.20 & 74.19 & 74.57 & 73.60 & 73.14 
\\

\fedmd {\cite{FedMD_NeurIPS19}} 
& 84.64 & 34.31 & 87.35 & \textbf{98.10} & 76.10 
\\

\fedmatch {\cite{FedMatch_ICLR21}} 
& 86.60 & 49.98 & 87.45 & 97.58 & 80.40
\\

\rhfl{} \cite{RHFL_CVPR22} 
& 75.06 & 55.19 & \textbf{88.49} & 97.23 & 78.99
\\

\hline

Our \fccl{} \cite{FCCL_CVPR22} 
& \underline{87.34} & \textbf{85.03} & \underline{87.89} & 93.98 & \underline{88.56} 
\\

Our \oursabbrv{} 
& \textbf{91.32} & \underline{84.70} & 87.41 & \underline{98.08} & \textbf{90.37} 

\end{tabular}}}
\vspace{-10pt}
\end{table}

\subsection{Diagnostic Experiments}
\label{sec:ablation}
{\color{majorblack}
For thoroughly assessing the efficacy of essential components of our approach, we perform an ablation study on \digits{} and \officeTO{} to investigate the effect of each essential component: \FCCM{} (\FCCMab{} \cref{sec:3.1}), \FISL{} (\FISLab{} \cref{sec:3.2}) and \FNTD{} (\FNTDab{} \cref{sec:3.3}). We firstly give a  quantitative result of inter- and intra-domains performance on these three components in \cref{table:ablation}. For the local updating phase without \FNTDab{} module, experiments remain same optimization objective as our conference version in \cref{eq:locafccl} because \FCCMab{} and \FISLab{} are designed for how to better learn from others in collaborative updating, rather than alleviate catastrophic forgetting. The first row refers to a baseline that only respectively trains on private data. Three crucial conclusions can be drawn. 
\textbf{First}, \FCCMab{} leads to significant performance improvements against the baseline across all the metrics on both scenarios.  This evidences that our cross-correlation learning strategy is able to produce generalizable performance. 
\textbf{Second},  we also observe compelling gains by incorporating \FISLab{} into the baseline. This proves  the importance of feature-level communication. Besides, combining \FCCMab{} and \FISLab{} achieves better performance, which supports our motivation of exploiting joint logits and feature levels communication rather than only focusing on logits output in model heterogeneity. 
\textbf{Third}, our full method leads to the best performance, by further leveraging \FNTDab{} in local updating phase, which suggests that these modules are complementary to each other, and confirms the effectiveness of our whole design. Besides, \oursabbrv{} maintains the robustness with diverse architecture selection, as illustrated in \cref{table:ablationwitharchitecture}. 
}

\begin{figure*}[t]
\centering
\subfigure[Epoch 1]{
\includegraphics[width=0.085\textwidth]{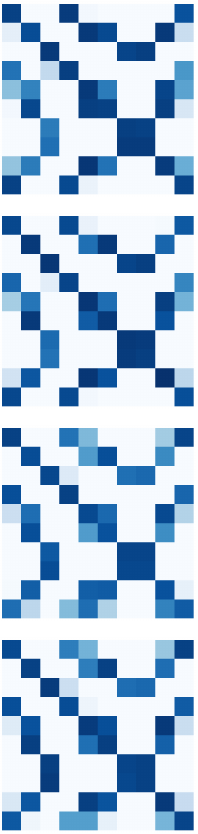}
\includegraphics[width=0.085\textwidth]{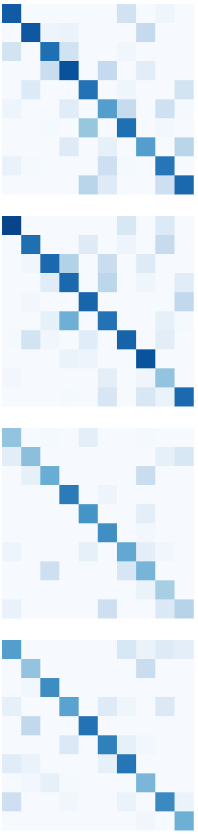}
}
\subfigure[Epoch 5]{
\includegraphics[width=0.085\textwidth]{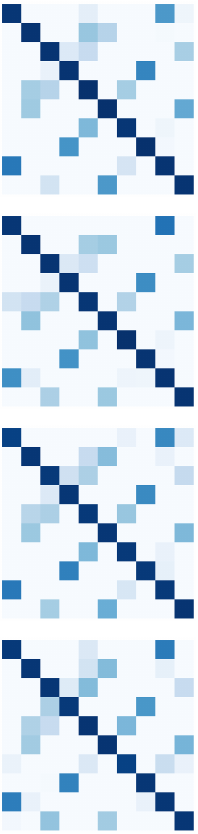}
\includegraphics[width=0.085\textwidth]{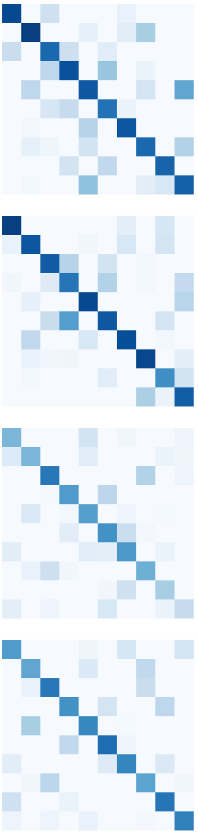}
}
\subfigure[Epoch 10]{
\includegraphics[width=0.085\textwidth]{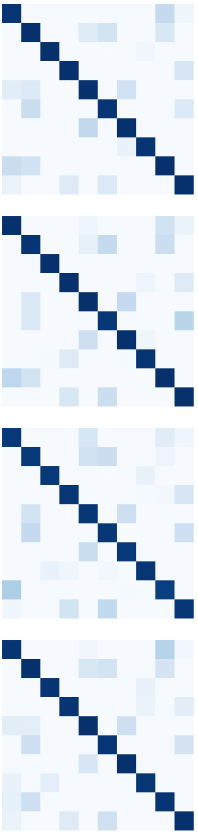}
\includegraphics[width=0.085\textwidth]{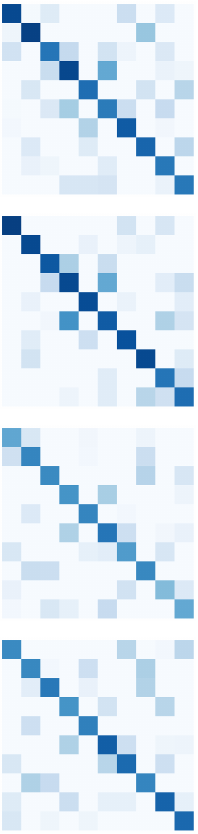}
}
\subfigure[Epoch 20]{
\includegraphics[width=0.085\textwidth]{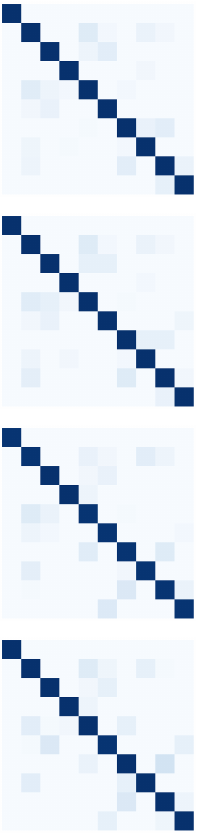}
\includegraphics[width=0.085\textwidth]{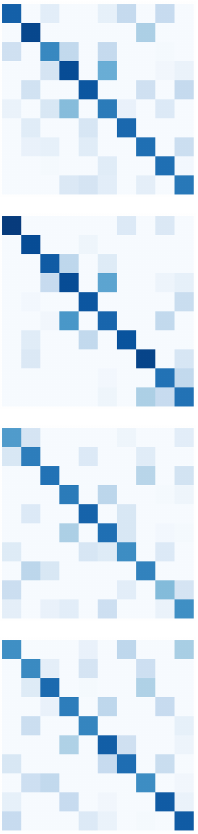}
}
\subfigure[Epoch 40]{
\includegraphics[width=0.085\textwidth]{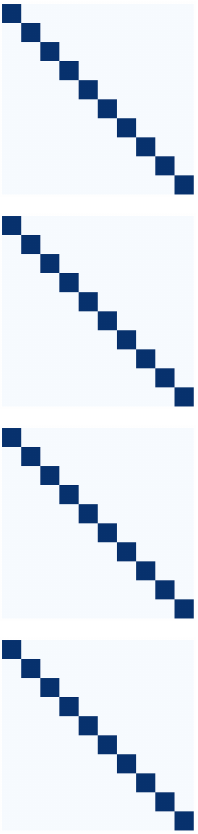}
\includegraphics[width=0.085\textwidth]{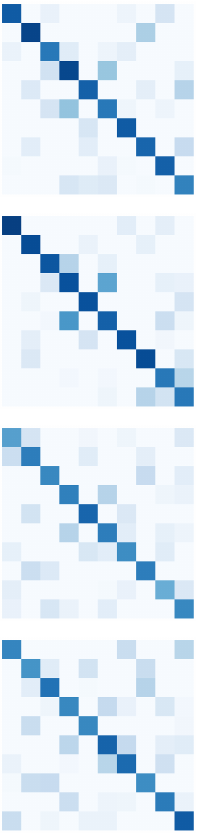}
}
\put(-520,150){\textbf{\mnist}}
\put(-520,105){\textbf{\usps}}
\put(-520,60){\textbf{\svhn}}
\put(-520,15){\textbf{\syn{}}}
\vspace{-10pt}
\caption{\textbf{Cross-correlation matrix visualization} for different domains in \digits{} scenario with \cifarhun{}. The matrix is $10 \times 10$. We visualize the cross-correlation matrix ($\mathcal{M}_i$ in \cref{eq:avgcrosscorr}) with others on unlabeled public data (\textbf{Left Column}). Besides, for each domain, we represent the $\mathcal{M}_i$ with others on respective private data (\textbf{Right Column}). The darker the color, the closer the $\mathcal{M}_i^{uv}$  is to $1$. Best viewed in color. See details in \cref{sec:ablation_fccm}.}
\vspace{-10pt}
\label{fig:fcclHeatMap}
\end{figure*}

\subsubsection{\FCCM{}}
\label{sec:ablation_fccm}
To prove \FCCM{} (\FCCMab{}) robustness and stability, we next evaluate the performance on different public data (\ie, \cifarhun{}, \tyimagenet{} and \fashionmnist{}). The \cref{fig:ab_fcc_inter} suggests that \FCCM{} achieves consistent performance in each domain. Moreover, it can be seen that it is more effective by the use of public data with rich categories (\tyimagenet) or simple details (\fashionmnist). 
Besides, the \cref{fig:fcclHeatMap} presents that \FCCMab{}{} achieves similar logits output among participants and minimizes the redundancy within the logits output, confirming that \FCCMab{} successfully enforces the correlation of same dimensions and decorrelation of different dimensions on both public and private data.

\begin{table}[t]
\centering
\caption{\textbf{Analysis of public data} diversity and data augmentation effect for \FISLab{} (\cref{sec:3.2}) in \digits{} and \officeTO{} scenarios. See details in \cref{sec:ablation_frsl}.}
	\label{table:public_with_frsl}
	\vspace{-10pt}
	\resizebox{\columnwidth}{!}{
		\setlength\tabcolsep{1.3pt}
		\renewcommand\arraystretch{1.3}
		\begin{tabular}{c||ccccIcIcccIc}
			\hline\thickhline
			\rowcolor{mygray}
			& \multicolumn{5}{cI}{\digits{}} & \multicolumn{4}{c}{\officeTO{}} \\ \cline{2-10}
			\rowcolor{mygray}
			\multirow{-2}{*}{Aug}& M$\rightarrow$ & U$\rightarrow$ & SV$\rightarrow$ & SY$\rightarrow$ & AVG & AM$\rightarrow$ & D$\rightarrow$ & W$\rightarrow$ & AVG \\
   			\hline\hline
    		\multicolumn{10}{l}{\fashionmnist{}} \\ \hline
            weak & 23.84 & 20.03 & 46.86 & 49.28 & \textbf{35.00}
            & 20.40 & 29.59 & 35.87 & \textbf{28.62}\\ 
            strong & 24.45 & 21.91 & 46.38 & 47.09 & 34.95
            & 19.00 & 28.37 & 35.42 & 27.59\\ 
            \hline\hline
            \multicolumn{10}{l}{\cifarhun{}} \\ \hline
            weak & 25.59 & 24.75 & 47.53 & 53.12 & \textbf{37.74} 
            & 20.02 & 28.04 & 36.51 & 28.19\\ 
            strong & 26.36 & 21.28 & 50.03 & 50.99 &  37.16 
            & 20.83 & 28.18 & 36.11 & \textbf{28.37}\\ 
            \hline\hline
            \multicolumn{10}{l}{\tyimagenet{}} \\ \hline
            weak & 27.60 & 27.08 & 45.66 & 51.94 & \textbf{38.07}  
            & 21.05 &  28.36 & 34.81 & 28.07 \\
            strong & 25.60 & 29.95 & 40.41 & 50.83 & 36.69
            & 21.12 & 30.43 & 34.14 & \textbf{28.56}\\
		\end{tabular}}
	\vspace{-10pt}
\end{table}

\begin{table}[t]
	\centering
	\caption{\textbf{Analysis of weighting hyper-parameter, $\omega$} for $\mathcal{L}_{\FISLab{}}$ in $ \mathcal{L}_{colla}$ (\cref{eq:col_loss}) in \digits{} and \officeTO{} with \cifarhun{}. See details in \cref{sec:ablation_frsl}.}
	\label{table:omegaabaltion}
	\vspace{-10pt}
	\resizebox{\columnwidth}{!}{
		\setlength\tabcolsep{1.3pt}
		\renewcommand\arraystretch{1.3}
		\begin{tabular}{c||ccccIcIcccIc}
			\hline\thickhline
			\rowcolor{mygray}
			$\omega$& \multicolumn{5}{cI}{\digits{}} & \multicolumn{4}{c}{\officeTO{}} \\ \cline{2-10}
			\rowcolor{mygray}
			\cref{eq:col_loss} & M$\rightarrow$ & U$\rightarrow$ & SV$\rightarrow$ & SY$\rightarrow$ & AVG & AM$\rightarrow$ & D$\rightarrow$ & W$\rightarrow$ & AVG \\
			\hline\hline
			$1$ & 25.54 & 21.95 & 50.22 & 52.08 & 37.44
			    & 25.35 & {31.22} & 37.57 & 31.38\\
			
			$2$ & {28.69} & 23.95 & 50.88 & 51.28 & 38.70
			    & {26.88} & {31.16} & {37.70} & 31.91\\

			$3$ & \textbf{30.97} & \textbf{25.90} & 51.12 & \textbf{53.84} & \textbf{40.45}
			    & 26.08 & \textbf{31.64} & \textbf{38.74} & \textbf{32.15}\\
      
			$4$ &  27.01 & {24.12} & \textbf{52.08} & 54.70 & 39.47
			    & \textbf{27.25} & 29.24 & 37.86 & 31.45\\
		\end{tabular}
	}
	\vspace{-15pt}
\end{table}

\begin{table*}[t]\small
\centering
\caption{\textbf{Comparison with state-of-the-art homogeneous federated learning methods under
model homogeneity} in \digits{} and \officeTO{} scenarios with \cifarhun{}. Both intra-domain discriminability ($\hookleftarrow$) and inter-domain generalization ($\rightarrow$) accuracies are reported. See details in \cref{sec:comparHomo}.}
\label{table:comparHomogeneity}
\vspace{-10pt}
{
\resizebox{0.9\textwidth}{!}{
\begin{tabular}{r||ccccIccIcccIcc}
\hline\thickhline
\rowcolor{mygray}
& \multicolumn{6}{cI}{\digits}&\multicolumn{5}{c}{\officeTO}\\
\cline{2-12}
\rowcolor{mygray}
\multirow{-2}{*}{Methods}  &  {M}$\rightarrow$  & {U}$\rightarrow$  & {SV}$\rightarrow$  & {SY}$\rightarrow$  & AVG  & $\triangle$ & {AM}$\rightarrow$   & {D}$\rightarrow$  & {W}$\rightarrow$  & AVG & $\triangle$\\

\hline\hline

\solo 
& 14.36 & 13.82 & 50.75 & 46.83 & 31.44 & -
& 22.00 & 32.50 & 36.16 & 30.22 & - \\

\fedavg{} {\cite{FedMD_NeurIPS19}} 
& 74.55 & 77.42 & 66.36 & 83.86 & 75.54 & +44.10
& 60.06 & 54.69 & 59.50 & 58.08 & +27.86 \\

\fedprox{} {\cite{FedProx_MLSys2020}} 
& \textbf{87.61} & \textbf{89.49} & \textbf{85.71} & {84.40} & \textbf{86.80} & \textbf{+55.36}
& \underline{64.74} & \textbf{61.32} & \underline{69.81} & \underline{65.29} & \underline{+35.07}\\

\fedmd{} {\cite{FedMD_NeurIPS19}} 
& 78.19 & \underline{81.29} & 66.53 & 80.81 & 76.70 & +45.26
& 66.09 & 59.05 & 66.08 & 63.74 & +33.52\\

\feddf{} {\cite{FedDF_NeurIPS20}} 
& 77.61 &  80.90 & 68.93 & 83.92 & 77.84 & +46.40
& 55.82 & 44.30 & 60.30 & 53.47 & +23.25 \\

\moon{} ($\tau=5$) {\cite{MOON_CVPR21}}
& 57.55 & 67.65 & 62.95 & 76.52 & 66.16 & +34.72
& 63.32 & \underline{58.72} & {64.42} & {62.15} & +31.93\\

\moon{} ($\tau=1$) {\cite{MOON_CVPR21}}
& 76.81 & 78.10 & \underline{72.99} & 83.50 & 77.85 & +46.41
& 64.76 & 62.20 & 67.53 & 64.83 & +34.61\\

\fedproc{} {\cite{FedProc_arXiv21}}
& 72.05 & 74.12 & 62.57 & 78.34 & 71.77 & +40.33
& 37.61 & 45.20 & 55.86 & 46.22 & + 16.00\\


\fedrs{}   {\cite{FedRS_KDD21}}
& 55.28 & 65.80 & 61.59 & 73.23 & 63.97 & +32.53
& 46.77 & 47.75 & 54.56 & 49.69 & +19.47\\


\hline
Our \fccl{} {\cite{FCCL_CVPR22}}
&  77.18 & 79.11 & 68.96 & \underline{85.26} & 77.62 & +46.18
& 55.39 & 46.16 & 61.60 & 54.38 & +24.16 \\

Our \oursabbrv{} 
& \underline{78.99} & 80.54 & {72.15} & \textbf{85.38} & \underline{79.26} & \underline{+47.82}
& \textbf{72.39} & {55.00} & \textbf{71.49} & \textbf{66.69} & \textbf{+36.47}\\

\hline\thickhline
\rowcolor{mygray}
& \multicolumn{6}{cI}{\digits{}}&\multicolumn{5}{c}{\officeTO{}}\\
\cline{2-12}
\rowcolor{mygray}
\multirow{-2}{*}{Methods} & M$\hookleftarrow$ & U$\hookleftarrow$ & SV$\hookleftarrow$ & SY$\hookleftarrow$ & AVG & $\triangle$ & AM$\hookleftarrow$ & D$\hookleftarrow$ & W$\hookleftarrow$ & AVG & $\triangle$ \\
\hline\hline

\solo 
& 82.09 & 77.15 & 90.10 & 98.08 & 86.85 & -
& 71.58 & 73.63 & 78.07 & 74.42 & -\\

\fedavg{} {\cite{FedMD_NeurIPS19}} 
& 92.50 & \textbf{91.48} & \underline{90.98} & 85.27 & 90.05 & \underline{+3.20}
& 71.25 & 77.58 & 78.20 & 75.67 & +1.25\\

\fedprox{} {\cite{FedProx_MLSys2020}} 
& \underline{92.44} & 90.36 & \textbf{92.08} & \textbf{96.20} & \textbf{92.77} & \textbf{+5.92}
& {71.58} & 76.24 & {80.71} & {76.17} & {+1.75}\\

\fedmd{} {\cite{FedMD_NeurIPS19}} 
& 90.55 & 88.82 & 90.79 & \underline{88.17} & {89.58} & +2.73
& 70.96 & 78.24 & 78.58 & 75.92 & +1.50\\

\feddf{} {\cite{FedDF_NeurIPS20}} 
& 91.08 &  89.70 & 90.81 & 87.51 & \underline{89.77} & \underline{+2.92}
& 70.95 & {77.84} & 78.99 & 75.92 & +1.50\\

\moon{} ($\tau=5$){\cite{MOON_CVPR21}}
& 87.58 & 84.46 &  87.70 & 70.28 &  82.50 & -4.35
& 67.51 & 70.68 & 74.01 & 70.73 & -3.69\\

\moon{} ($\tau=1$) {\cite{MOON_CVPR21}} 
& 91.63 & 89.02 & 90.91 & 77.27 & 87.20 & +1.35
& 69.00 & 74.90 & 76.61 & 73.50 & -0.92\\

\fedproc{}  {\cite{FedProc_arXiv21}} 
& 85.91 & 83.53 & 89.01 & 76.62 & 83.76 & -3.09
& 69.38 & 75.24 & 78.95 & 74.52 & +0.10\\


\fedrs{}  {\cite{FedRS_KDD21}}
& 82.27 & 85.91 & 86.79 & 71.65 & 81.65 & -5.20
& 70.86 & 75.63 & 79.22 & 75.23 & +0.81\\


\hline

Our \fccl{} {\cite{FCCL_CVPR22}}
& 90.71 & 90.61 & 90.92 &  83.40 & 88.91 & +2.06 
& \textbf{72.34} & 77.31 & \underline{81.68} & \underline{77.11} & \underline{+2.69}\\

Our \oursabbrv{} 
& \textbf{93.74} & \underline{90.78} & 90.71 & 80.68 & 88.97 & +2.12
& \underline{71.65} & \underline{78.58} & \underline{82.73} & \textbf{77.65} & \textbf{+3.23}
\end{tabular}}}
\vspace{-10pt}
\end{table*}

\begin{table}[t]
\centering
\caption{\color{black}{\textbf{Compare average inter-domain generalization performance of \FNTDab{} with EWC} {\cite{EWC_PNAS17}} in different hyper-parameter: $\lambda$ on \digits{} (\textbf{Left}) and \officeTO{} (\textbf{Right}) scenarios. See details in \cref{sec:ablation_fntd}.}}
\label{tab:comparewithewc}
\vspace{-10pt}
\centering
{
\resizebox{\linewidth}{!}{
\setlength\tabcolsep{2.0pt}
\begin{tabular}{c||ccccIc||cccc}
\hline \thickhline
\rowcolor{mygray}
& \multicolumn{4}{cI}{EWC ($\lambda$)} 
&
& \multicolumn{4}{c}{EWC ($\lambda$)} \\
\cline{2-5}
\cline{7-10}
\rowcolor{mygray}
\multirow{-2}{*}{Ours} 
& 0.01 & 0.1 & 0.7 & 1.0
& \multirow{-2}{*}{Ours} 
& 0.01 & 0.1 & 0.7& 1.0\\
\hline \hline
\textbf{50.49} & 28.50 & 28.38 & \underline{29.08} & 28.70 & 
\textbf{37.21} & \underline{29.58} & 28.59 & 29.04 & 29.18
\end{tabular}}}
\vspace{-10pt}
\end{table}

\begin{table}[t]
\centering
\caption{\color{black}{\textbf{Ablation of incorporating KD} with \FNTDab{} in \digits{} and \officeTO{} scenarios. This ratio represents the proportion of KD in the local training process. “0” means purely leveraging \FNTDab{} in the local updating. Please see details in \cref{sec:ablation_fntd}.}}
\label{table:ablationwithKDNKD}
\vspace{-10pt}
\resizebox{\columnwidth}{!}{
\setlength\tabcolsep{1.1pt}
\renewcommand\arraystretch{1.3}
\begin{tabular}{r||ccccIcIccccIc}
\hline\thickhline
\rowcolor{mygray}
&  \multicolumn{10}{c}{\digits{}} \\ 
\cline{2-11}
\rowcolor{mygray}
\multirow{-2}{*}{w KD}  & 
M$\rightarrow$ & U$\rightarrow$ & SV$\rightarrow$ & SY$\rightarrow$ & AVG & 
M$\hookleftarrow$ & U$\hookleftarrow$ & SV$\hookleftarrow$ & SY$\hookleftarrow$ & AVG \\
\hline\hline
30\% &
24.25 & 25.39 & 45.55 & 56.54 & 37.93 &
87.79 & 85.10 & 86.70 & 97.12 & 89.17 \\ 
10\% &
26.66 & 33.61 & 48.70 & 56.56 & 41.38 &
89.54 & 88.59 & 87.32 & 97.07 & \textbf{90.63} \\ 
\hline\hline
0&
{40.13} & {50.53} & {48.31} & {63.00} & \textbf{50.49} & {90.54} & {86.92} & {87.11} & {97.72} & {90.57}
\end{tabular}}
\vspace{-10pt}
\end{table}

\begin{table}[t]
\centering
\caption{\color{majorblack}{\textbf{Ablation of different network selection} in the \digits{} scenario with \cifarhun{} dataset. The default architecture is in \cref{table:custmodel}. See \cref{sec:ablation}.}}
\label{table:ablationwitharchitecture}
\vspace{-10pt}
\resizebox{\columnwidth}{!}{
\setlength\tabcolsep{1.1pt}
\renewcommand\arraystretch{1.3}
\begin{tabular}{r||ccccIcIccccIc}
\hline\thickhline
\rowcolor{mygray}
&  \multicolumn{10}{c}{\digits{}} \\ 
\cline{2-11}
\rowcolor{mygray}
\multirow{-2}{*}{Methods}  & 
M$\rightarrow$ & U$\rightarrow$ & SV$\rightarrow$ & SY$\rightarrow$ & AVG & 
M$\hookleftarrow$ & U$\hookleftarrow$ & SV$\hookleftarrow$ & SY$\hookleftarrow$ & AVG \\
\hline\hline
\multicolumn{11}{l}{
M: \resnetten{} U: \resnettwelve{} SV: \efficientnet{} SY: \simplecnn
} \\ \hline
\fedmd{} &
20.83 & 16.92 & 48.73 & 31.45 & 29.48 &
85.00 & 67.02 & 86.49 & 87.38 & 81.47 \\ 
\fccl{} &
22.09 & 16.71 & 50.49 & 37.63 & 31.73 &
86.98 & 82.33 & 88.26 & 84.12 & 85.42 \\ 
\oursabbrv{}&
\textbf{37.13} & \textbf{45.97} & \textbf{49.12} & \textbf{52.99} & \textbf{46.30} &
\textbf{91.79} & \textbf{88.24} & \textbf{87.57} & \textbf{86.17} & \textbf{88.44} \\ 
\hline
\multicolumn{11}{l}{
M: \resneteighteen{} U: \resnettwelve{} SV: \efficientnet{} SY: \mobilenet{}
} \\
\hline
\fedmd{} &
19.27 & 14.58 & 43.74 & 46.18 & 30.94 &
90.20 & 55.12 & 87.16 & 97.80 & 82.57 \\ 
\fccl{} &
24.74 & 16.57 & 50.63 & 49.35 & 35.32 &
89.98 & 83.76 & 87.22 & 97.28 & 89.56 \\ 
\oursabbrv{} &
\textbf{56.71} & \textbf{45.07} & \textbf{53.03} & \textbf{61.05} & \textbf{53.96} &
\textbf{93.23} & \textbf{87.31} & \textbf{87.94} & \textbf{96.92} & \textbf{91.35} \\ 
\end{tabular}}
\vspace{-10pt}
\end{table}

\begin{table}[t]
\centering
\caption{\color{majorblack}{\textbf{Catastrophic forgetting} effect with different local epochs in the \digits{} scenario with \cifarhun{} dataset. w/o \FNTDab{} denotes without \FNTDab{} module in \cref{eq:localobj}. Refer to \cref{sec:ablation_fntd} for details.}}
\label{table:ablationCF}
\vspace{-10pt}
\resizebox{\columnwidth}{!}{
\setlength\tabcolsep{1.1pt}
\renewcommand\arraystretch{1.3}
\begin{tabular}{r||ccccIcIccccIc}
\hline\thickhline
\rowcolor{mygray}
Local& \multicolumn{5}{cI}{w/o \FNTDab{}} & \multicolumn{5}{c}{w/ \FNTDab{}}  \\ 
\cline{2-11}
\rowcolor{mygray}
Epoch & 
M$\rightarrow$ & U$\rightarrow$ & SV$\rightarrow$ & SY$\rightarrow$  & AVG &  
M$\rightarrow$ & U$\rightarrow$ & SV$\rightarrow$ & SY$\rightarrow$ & AVG \\
\hline\hline
10 &         
18.56 & 15.54 & 45.71 & 47.37 & 31.79 & 
21.81 & 16.09 & 48.54 & 49.78 & \textbf{34.05}
\\ 
20 &
17.50 & 15.35 & 48.32 & 54.67 & 33.96 &
40.13 & 50.53 & 48.31 & 63.00 & \textbf{50.49}\\
30 &
17.15 & 16.23 & 47.93 & 50.25 & 32.89 & 
33.35 & 40.69 & 49.82 & 57.49 & \textbf{45.33}\\ 
\end{tabular}}
\vspace{-10pt}
\end{table}

\begin{figure*}[t]
\noindent \begin{minipage}[]{0.6\textwidth}
\centering
\subfigure[M$\rightarrow$M]{\includegraphics[width=0.24\linewidth]{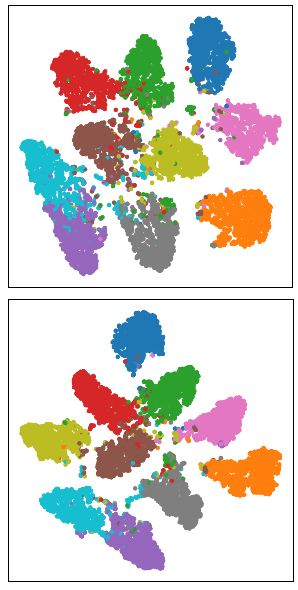}}
\subfigure[M$\rightarrow$U]{\includegraphics[width=0.24\linewidth]{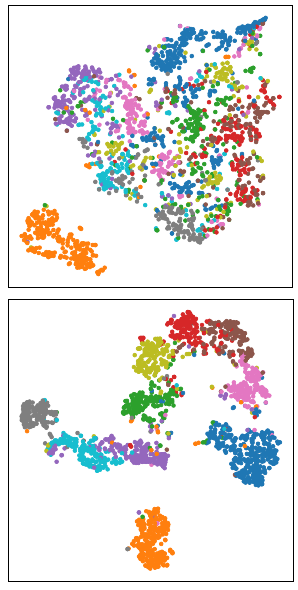}}
\subfigure[U$\rightarrow$U]{\includegraphics[width=0.24\linewidth]{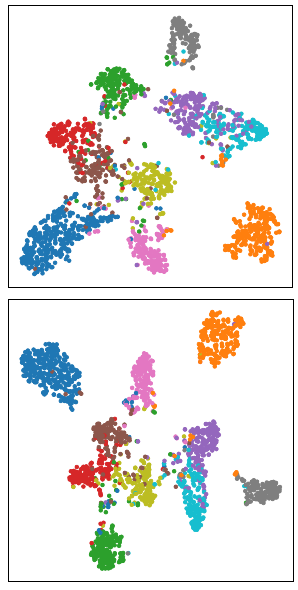}}
\subfigure[U$\rightarrow$M]{\includegraphics[width=0.24\linewidth]{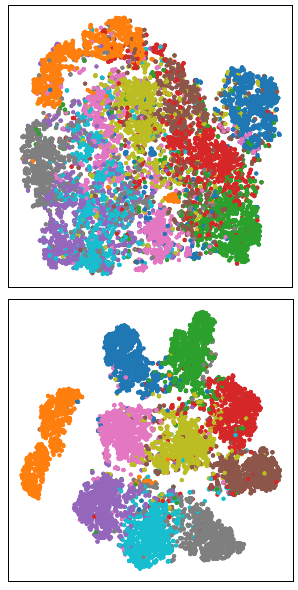}}
\put(0,105){\textbf{ $\theta^e$}}
\put(0,35){\textbf{ $\theta^{e-1}$}}
\end{minipage}
\hfill
\noindent\begin{minipage}[]{0.35\textwidth}
\centering
\small{
	\resizebox{\linewidth}{!}{
		\setlength\tabcolsep{1.3pt}
		\renewcommand\arraystretch{1.4}
		\begin{tabular}{c||ccccIc}
			\hline\thickhline
			\rowcolor{mygray}
			$\mathcal{T}$& \multicolumn{5}{c}{\digits{}}  \\ \cline{2-6}
			\rowcolor{mygray}
			\cref{eq:fntd} & M$\rightarrow$ & U$\rightarrow$ & SV$\rightarrow$ & SY$\rightarrow$ & AVG \\
			\hline\hline
			$\theta^e$ & 16.08 & 20.25 & 34.75 & 45.95 & 29.25 \\
			$\theta^{e-1}$ & \textbf{40.13} & \textbf{50.53} & \textbf{48.31} & \textbf{63.00} &  \textbf{50.49} \\

			\hline \thickhline
			\rowcolor{mygray}
			$\mathcal{T}$& \multicolumn{5}{c}{\digits{}}  \\ \cline{2-6}
			\rowcolor{mygray}
			\cref{eq:fntd} & M$\hookleftarrow$ & U$\hookleftarrow$ & SV$\hookleftarrow$ & SY$\hookleftarrow$ & AVG \\
			 			\hline\hline
 			$\theta^e$ & 88.31 & 73.11 & 82.95 & 95.58 & 84.98 \\

			$\theta^{e-1}$ & \textbf{90.54} & \textbf{86.92} & \textbf{87.11} & \textbf{97.72} & 
		    \textbf{90.57}\\
		\end{tabular}
	}}
\end{minipage}
\vspace{-10pt}
\caption{\textbf{Analysis of teacher}, $\mathcal{T}$ in \FNTDab{} (\cref{eq:fntd}) for \digits{} scenario with \cifarhun{}. We visualize the learned feature on intra-domain (\ie, M $\rightarrow$ M) and inter-domain (\ie, M $\rightarrow$ U) with $\theta^{e-1}$ (\textbf{Bottom Row}) and $\theta^e$ (\textbf{Top Row}). Features are colored based on labels. Refer to \cref{sec:ablation_fntd}.}
\label{fig:teacherinFNTD}
\vspace{-10pt}
\end{figure*}

\begin{figure*}[t]
\noindent \begin{minipage}[]{0.26\textwidth}
\centering
\includegraphics[width=\linewidth]{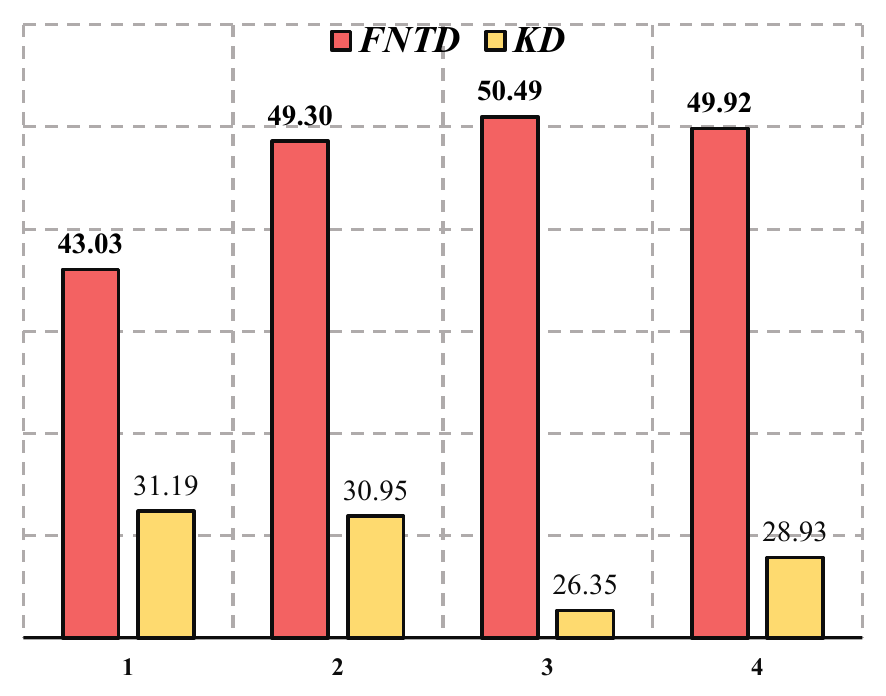}
\end{minipage}
\hfill
\noindent \begin{minipage}[]{0.22\textwidth}
\centering
\includegraphics[width=\linewidth]{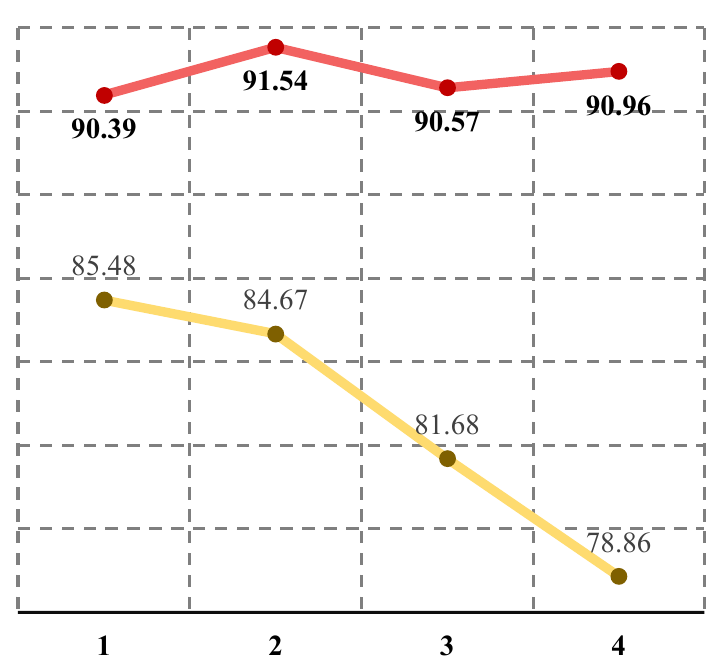}
\end{minipage}
\noindent \begin{minipage}[]{0.26\textwidth}
\centering
\includegraphics[width=\linewidth]{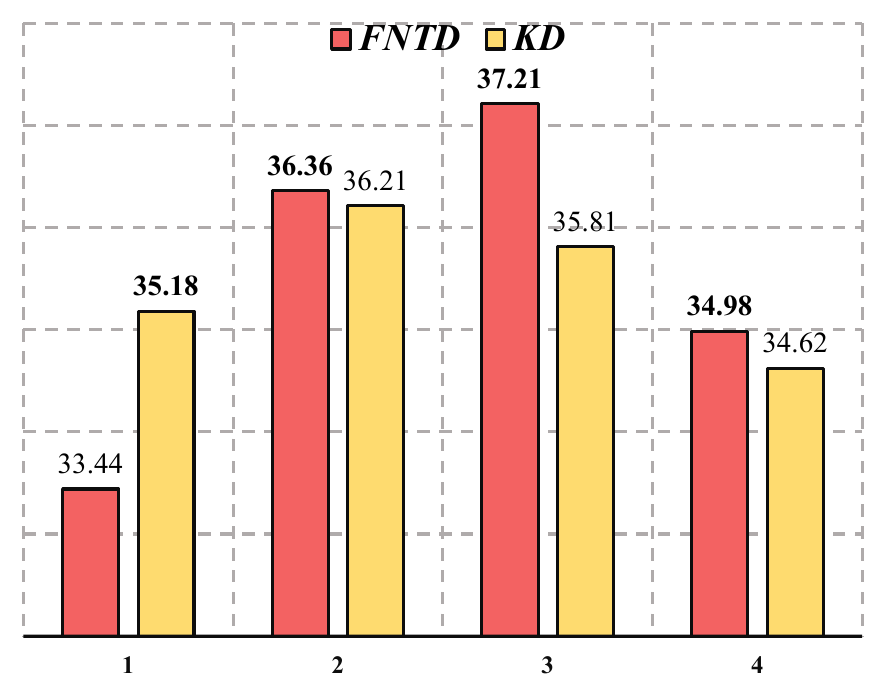}
\end{minipage}
\hfill
\noindent \begin{minipage}[]{0.22\textwidth}
\centering
\includegraphics[width=\linewidth]{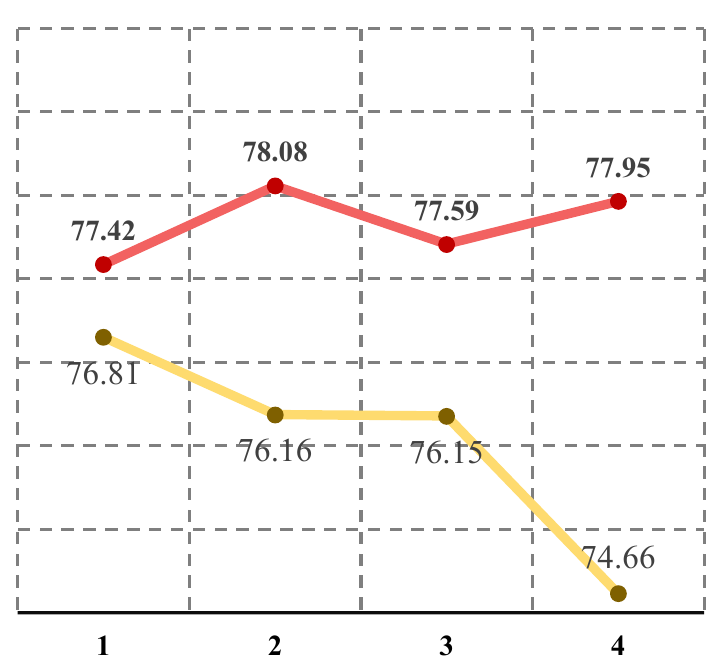}
\end{minipage}
\vspace{-5pt}
\caption{
{\color{majorblack}{\textbf{Comparison of \FNTDab{}} (\cref{eq:fntd}) and \textbf{KD} (\cref{eq:kd_ori_form}) with different \textbf{temperature hyper-parameter}, $\tau$ in \cref{eq:kd_component} for inter-domain performance (\textbf{Histogram}) and intra-domain performance (\textbf{Line}) in \digits{} and \officeTO{} scenarios with \cifarhun{}. See details in \cref{sec:ablation_fntd}.}
}}
\vspace{-15pt}
\label{fig:ab_fntd}
\end{figure*}

\subsubsection{\FISL}
\label{sec:ablation_frsl}
We next examine the design of our \FISL{} (\FISLab{}). \FISLab{} aims to reach similarity distribution alignment on unlabeled public data. Thus, we analyze the impact of  diversity and augmentation on public data in \cref{table:public_with_frsl}. The results reveal that for simple scenarios, \ie, \digits{} (10 categories), leveraging weak augmentation is more beneficial, which is contrary for complicated scenarios \ie, \officeTO{} (31 categories). Besides, different scenarios would benefit from the diversity of public data. Specifically, in \digits{} scenario, the average inter-domain performance increase from $37.74\%$ to $38.07\%$ when public data change from \cifarhun{} to \tyimagenet{} with weak data augmentation. The details of weak and strong data augmentation are listed in \cref{sec:ExperSetup}.
{\color{black}{
It shows that the proposed module \FISLab{} is robust under different data augmentations and shows comparable performance under different scenarios.}}
Then, in \cref{table:omegaabaltion}, we analyze the impact of the weighting hyper-parameter, $\omega$ for $\mathcal{L}_{\FISLab{}}$ in $ \mathcal{L}_{colla}$ (\cref{eq:col_loss}). As seen, the inter-domain performance progressively improves as $\omega$ is increased, and the gain becomes marginal when $\omega\!=\!3$. Hence, we select the $\omega\!=\!3$ by default in different scenarios.

\subsubsection{\FNTD}
\label{sec:ablation_fntd}
As shown in \cref{table:ablationCF}, without inter-domain knowledge regularization, local model would present serious catastrophic forgetting phenomenon on the inter-domain performance, demonstrating the importance of incorporating inter-domain knowledge in the local updating stage. We then investigate the effectiveness of \FNTD{} (\FNTDab{}). In \cref{fig:ab_fntd}, the \ding{55} refers to the typical knowledge distillation (KD) in {\cref{eq:kd_ori_form}}. Our \FNTDab{} outperforms the typical KD by avoiding optimization  conflict. The gains become larger in the relatively easy scenario, \ie, \digits{} because the  objective is easier contradictory under the supervision of  private data label information and more 'confident' previous model. Besides, we assess the impact of the temperature hyper-parameter, $\tau$  (\cref{eq:kd_component}) in $\mathcal{L}_{\FNTDab{}} $ (\cref{eq:fntd}). We find that the suitable temperature is $3$.  We believe this happens because a small temperature obstacles more information of non-targets to be distilled and a large temperature blindly softens logits  distribution, which falls back to a uniform distribution and is not informative, corroborating relevant observations reported in {\cite{KD_arXiv15,RNNwtKD_ICASSP16,RethinkSoftLabelforKD_ICLR21}}. Hence, we apply temperature $\tau\!=\!3$ by default. Finally, we study the selection of the teacher model in \FNTDab{} ({\cref{eq:fntd}}). We give a comparison of using different models: $\theta^{e-1}$ (after local updating) and $\theta^e$ (after collaborative updating) as teacher model in local updating. We draw t-SNE {\cite{tSNE_JMLR08}} visualization of data embeddings on the testing data and inter-domain performance by the end of federated learning in \cref{fig:teacherinFNTD}. This proves that $\theta^{e-1}$ is better than $\theta^{e}$ to maintain inter- and intra-domains knowledge in local updating simultaneously.
{\color{black}{
We further replace the proposed module \FNTDab{} with the EWC penalty \cite{EWC_PNAS17} in \cref{tab:comparewithewc}. It reveals that ours performs better than the existing relative methods in handling the catastrophic forgetting problem. 
Besides, we consider investigating the effectiveness of target distillation in the first half of local updating rounds in \cref{table:ablationwithKDNKD}. It reveals that purely leveraging \FNTDab{} achieves better performance. We argue that the confidence in the target class is uncontrollable and thus results in the underlying optimization conflict. Besides, defining the proportion of utilization rounds would bring additional hyper-parameter.}}

\subsection{Model Homogeneity Analysis}
\label{sec:comparHomo}
{\color{majorblack}
We further compare \oursabbrv{} with other methods under model homogeneous setting. We set the global shared model as \resnettwelve{} and add the model parameter averaging operation between \textbf{collaborative updating} and \textbf{local updating}. The \cref{table:comparHomogeneity} presents both inter- and intra-domains performance on \digits{} and \officeTO{} scenarios with \cifarhun{}. Although \fedprox{} achieve competitive performance in certain simple scenarios, \ie, \digits{}, it requires parameter element-wise regularization, which incurs heavy computation cost and limits the generalization ability \cite{FedAlign_CVPR22} in the challenging scenarios with the large domain shift such as \officeTO{}. In contrast, our method relies on the logits output, which remains a static target without recycle updating requirements and does not largely increase the computational burden with different network backbones.
Therefore, considering the computation cost and the underlying large domain shift in heterogeneous federated learning scenarios, FCCL+ emerges as a superior candidate compared to FedProx.
}

\section{Conclusion}
In this paper, we present \oursabbrv{}, a novel and effective solution for heterogeneous federated learning.  \oursabbrv{} exploit both logits and feature  levels knowledge communication by constructing cross-correlation matrix and measuring instance similarity distribution on unlabeled public data, effectively overcoming communication barrier and acquiring generalizable ability in heterogeneous federated learning. Besides, for alleviating catastrophic forgetting problem, we disentangle typical knowledge distillation to avoid optimization conflict and better preserve inter-domain knowledge. Moreover, we maintain the supervision of label signal to provide strong intra-domain constraint, boosting both inter- and intra-domains performance.  We experimentally show that \oursabbrv{} performs favorably with many existing related methods on four different scenarios. We wish this work to pave the way for future research on heterogeneous federated learning as it is fully reproducible and includes:

\begin{itemize} 
	\item \oursabbrv{}, a strong baseline that outperforms others while maintaining limited communication burden on server and friendly computation cost for participants.
	\item A clear and extensive benchmark comparison of the state-of-the-art on multiple domain shift scenarios.
\end{itemize}

\ifCLASSOPTIONcaptionsoff
  \newpage
\fi


%
{\small
\bibliographystyle{IEEEtran}

\bibliography{egbib}
}



\vfill


\end{document}